\pdfoutput=1
\documentclass[letterpaper]{article} 
\usepackage[preprint]{aaai2027}
\usepackage[hyphens]{url}  
\usepackage{graphicx} 
\usepackage{natbib}  
\usepackage{caption} 
\usepackage{algorithm}
\usepackage{algorithmic}
\usepackage{multirow}
\usepackage{xspace}
\usepackage{amsmath,amssymb,amsthm}
\usepackage{newfloat}
\usepackage{listings}
\usepackage{enumitem}
\usepackage{array}
\usepackage{colortbl}
\usepackage[most]{tcolorbox}
\usepackage{arydshln}   

\definecolor{lightblue}{RGB}{220, 235, 250}  
\definecolor{modelgray}{RGB}{240, 240, 240}
\DeclareCaptionStyle{ruled}{labelfont=normalfont,labelsep=colon,strut=off} 
\floatstyle{ruled}
\newfloat{listing}{tb}{lst}{}
\floatname{listing}{Listing}

\usepackage{booktabs}

\title{SearchAuditor: Auditing and Attributing Failures \\ in Long-Horizon Search Agents}
\author{
    Zhixiang Liang\equalcontrib\textsuperscript{\rm 1,\rm 2},
    Yifei Liu\equalcontrib\textsuperscript{\rm 2},
    Yidan Huang\textsuperscript{\rm 2},
    Haozhe Zhao\textsuperscript{\rm 1},\\
    Beichen Huang\textsuperscript{\rm 1},
    Jiaqi Wang\textsuperscript{\rm 2},
    Nan Duan\textsuperscript{\rm 2},
    Qiong Cao\textsuperscript{\rm 2}\corresponding
}
\affiliations{
    \textsuperscript{\rm 1}University of Illinois Urbana-Champaign\quad
    \textsuperscript{\rm 2}Joy Future Academy, JD
}

\begin{document}

\maketitle

\begin{abstract}


Deep search agents tackle challenging questions through long-horizon web interactions, a process that is both complex and fragile: small reasoning errors may propagate through long, noisy trajectories into fluent but incorrect answers. Diagnosing such failures is difficult, requiring the manual inspection of extremely long execution traces, which could be beyond human capacity. We therefore introduce SearchAuditBench, a benchmark that evaluates whether LLM auditors can localize, attribute, and repair these failures, thereby reducing the human burden. SearchAuditBench comprises 1,243 failed trajectories, averaging 73.1 messages and 65.1K tokens, collected from eight open-weight models on five deep-search benchmarks, each expert-annotated with the critical error step, a search-specific root cause, and a reference repair with grading rubrics. We further propose SearchAuditor, a multi-perspective auditing framework that effectively localizes, attributes, and repairs search-agent failures through evidence-grounded adjudication. Experimental results show that even the strongest baseline, when powered by a frontier model like GPT-5.5, attains only a 26.6\% end-to-end pass rate. In contrast, our SearchAuditor consistently outperforms all baselines across different frontier models, achieving an end-to-end pass rate of 32.3\%, and resuming failed runs with its repairs enables agents to better recover from errors.\textsuperscript{1}

\end{abstract}

\section{1. \ Introduction}

The rapid progress in large language models (LLMs) has enabled increasingly capable long-horizon agents \cite{singh2026openaigpt5card, glm5team2026glm5vibecodingagentic}. Among them, search agents have emerged as a prominent application \cite{nakano2021webgpt,wei2025browsecompsimplechallengingbenchmark}. Given an information-seeking question, the agent iteratively formulates queries, browses and reads web pages, and maintains candidate answers before committing to a final response \cite{li2025websailor}. Despite this, search agents still remain brittle over extended search processes, where small reasoning errors compound across steps and lead to confidently wrong answers \cite{zhan2026deepresearchagentfails}. Understanding these failures is essential to improving agents, yet it requires developers to manually inspect long execution traces to determine what went wrong, a process that scales poorly with trajectory length and volume \cite{cemri2026why}. Automatically auditing and attributing \emph{where} and \emph{how} a long-horizon search agent fails remains an under-explored problem.

\begin{figure}
    \centering
    \includegraphics[width=1\linewidth]{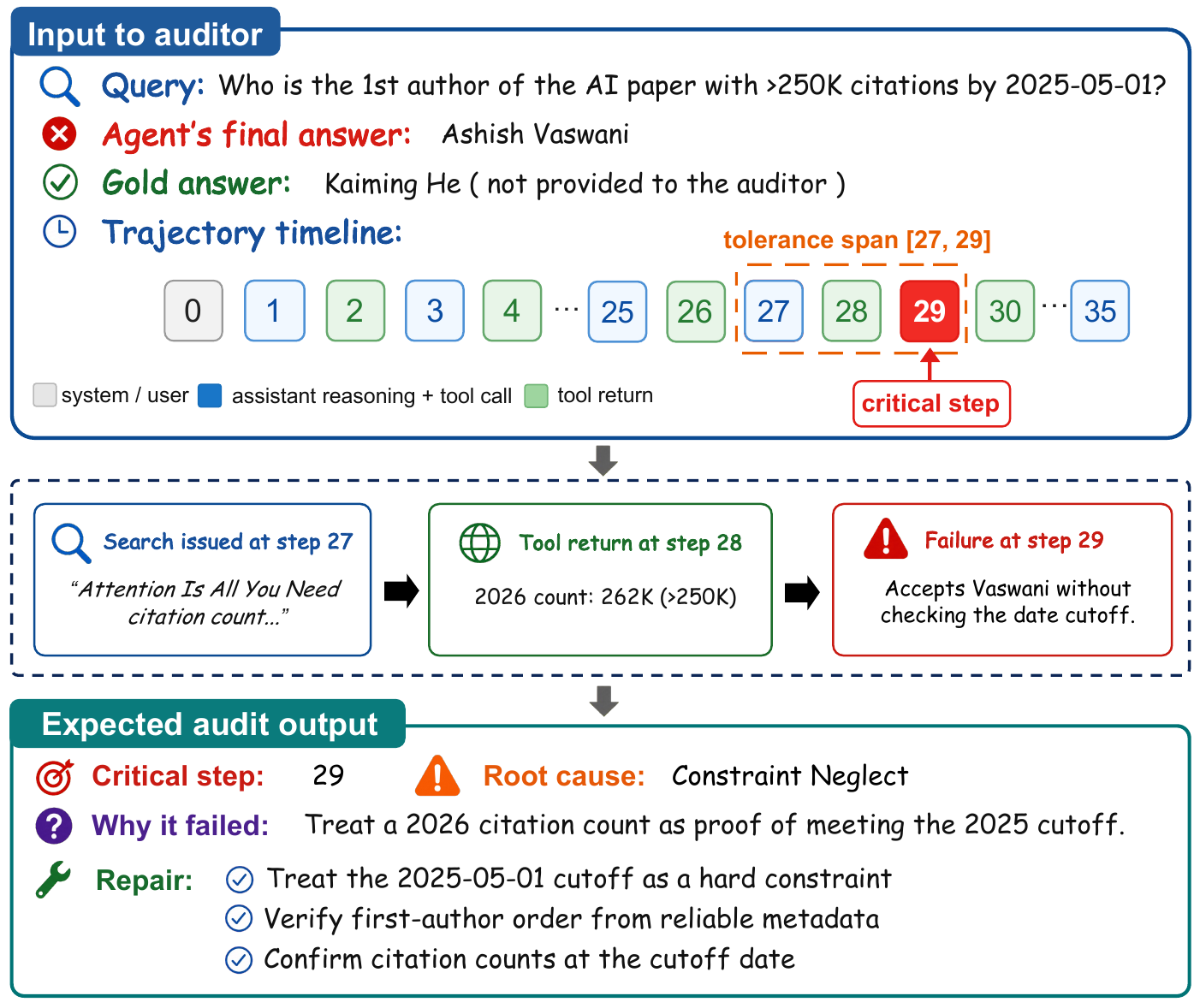}
    \caption{Illustration of the trajectory auditing task: given a query, the agent's wrong answer, and a failed trajectory, the auditor must localize the critical step, attribute a root cause, and prescribe a repair.}
    \label{fig:intro_figure}
\end{figure}

Recent work increasingly studies failure diagnosis over agent trajectories, including failure attribution in multi-agent systems \cite{zhang2025which, zhang2025agentracerinducingfailurellm}, constraint-based diagnosis from execution logs \cite{barke2026agentrxdiagnosingaiagent}, dependency-guided fault localization \cite{rafi2026falat}, and domain-specific tracing for memory and code agents \cite{deng2026memtrace, li2026codetracertraceableagentstates}. However, these efforts rarely target long-horizon search agents, whose trajectories pose distinctive diagnostic challenges.
(1) \textbf{Scale and noise}: a trajectory interleaves dozens of searches and page visits with lengthy observations, burying the critical error among many benign steps \cite{gou2026mind2web, wei2025browsecompsimplechallengingbenchmark}. (2) \textbf{Lack of verifiable structure}: unlike code agents checkable against executable tests or workflow agents with schema-defined invariants, open-web search offers no oracle; a step's correctness depends on evidence chains over noisy, ephemeral web content \cite{krishna-etal-2025-fact}. (3) \textbf{Silently propagating failures}: search agents rarely crash but fail subtly, trusting unverified sources, overlooking key evidence, or mismanaging candidates, and one early misstep propagates through subsequent reasoning while the final answer still appears fluent \cite{chen-etal-2026-beyond-single}.
Compounding these challenges, the community lacks a public, expert-annotated corpus of failed search trajectories labeled with critical error steps and root causes, which is essential for analyzing failure patterns and rigorously evaluating automated auditors.

To address this gap, we present a systematic study of auditing and attributing failures in long-horizon search agents, with three contributions. First, we construct \textbf{SearchAuditBench}, an expert-annotated benchmark of 1,243 failed trajectories, collected by running eight open-weight agents on five deep-search benchmarks under a unified scaffold. Each instance is annotated, via LLM-assisted screening and expert annotation, with (i) a critical error step and its tolerance span, (ii) a root cause from a six-way search-specific taxonomy, and (iii) a reference repair with atomic rubrics, which together define the three auditing tasks (Figure 1).
The corpus also reveals how search agents fail: 45.9\% of critical errors occur in the final third of a trajectory, and the dominant root cause, Candidate Mismanagement, accounts for 27.2\% of failures. Second, we benchmark LLM auditors on these tasks and find auditing highly challenging: even the strongest baseline, powered by a frontier model like GPT-5.5, reaches only a 26.6\% end-to-end fully passed score.
Third, we propose \textbf{SearchAuditor}, which runs three complementary audit branches in parallel, adjudicates their reports over a compact evidence-grounded view of the trajectory, and synthesizes process-level repair directives conditioned on the adjudicated diagnosis. It consistently outperforms all baselines across three frontier backbones, achieving an end-to-end pass rate of 32.3\%, and its repairs fix 17.4\% of Kimi-K2.6's failed runs on LiveBrowseComp \cite{fan2026livebrowsecompsearchagentssearching}, lifting accuracy from 34.0\% to 45.1\%.



\section{2. \ Related Work}

Recent work moves agent evaluation beyond final-answer correctness toward explaining where and why a trajectory fails. For single-agent trajectories, TRAIL \cite{deshpande2025trailtracereasoningagentic} localizes errors to trace spans under a taxonomy of reasoning, execution, and planning errors, while AgentRx \cite{barke2026agentrxdiagnosingaiagent} diagnoses critical errors via constraints from logs and tool specifications. In multi-agent systems, Who\&When \cite{zhang2025which} attributes failures to the responsible agent and its decisive step, and AgenTracer \cite{zhang2025agentracerinducingfailurellm} scales supervision via counterfactual replay and fault injection. Domain-specific tracers further target agent memory \cite{deng2026memtrace}, coding-agent states \cite{wang2026trajauditautomatedfailurediagnosis}, and CLI trajectories \cite{zhao2026failureprocessanatomycli}. 
However, none targets long-horizon search agents, whose trajectories are far longer and noisier and lack the verifiable oracles these methods rely on. We therefore introduce SearchAuditBench, an expert-annotated benchmark of 1,243 failed deep-search trajectories (compared with prior benchmarks in Table \ref{tab:failure_attribution_benchmarks}), and SearchAuditor, a multi-perspective auditing framework with evidence-grounded adjudication.

\begin{table}[t]
  \centering
  \setlength{\tabcolsep}{2.4pt}
  \renewcommand{\arraystretch}{1.08}
  \scriptsize

  \resizebox{\columnwidth}{!}{%
    \begin{tabular}{@{}lccccc@{}}
      \toprule
      \multirow{2}{*}{Benchmark} &
      \multirow{2}{*}{Domain} &
      \multicolumn{3}{c}{Scale} &
      \multirow{2}{*}{\shortstack[c]{Avg.\ trace \\ length}} \\
      \cmidrule(lr){3-5}
      & & \#Trace & \#Bench & \#Model & \\
      \midrule
      Who\&When  & General        & 184 & 2 & 1 & 22.2 msg. \\
      AgentRx    & General        & 115 & 3 & 2 & 42.2 steps \\
      TRAIL      & General / Code & 148 & 2 & 2 & 13.4 spans \\
      MemTrace   & Memory         & 160 & 3 & 1 & n/a \\
      TrajAudit     & Code           & 102 & 3 & 6 & 51.7 steps \\
      \midrule
      \textbf{SearchAuditBench} & \textbf{Search} & \textbf{1,243} &
      \textbf{5} & \textbf{8} & \textbf{73.1 msg.} \\
      \bottomrule
    \end{tabular}%
  }

  \caption{Comparison with prior trajectory-level failure attribution benchmarks.}
  \label{tab:failure_attribution_benchmarks}
\end{table}

\begin{figure*}[ht]
    \centering
    \includegraphics[width=\textwidth]{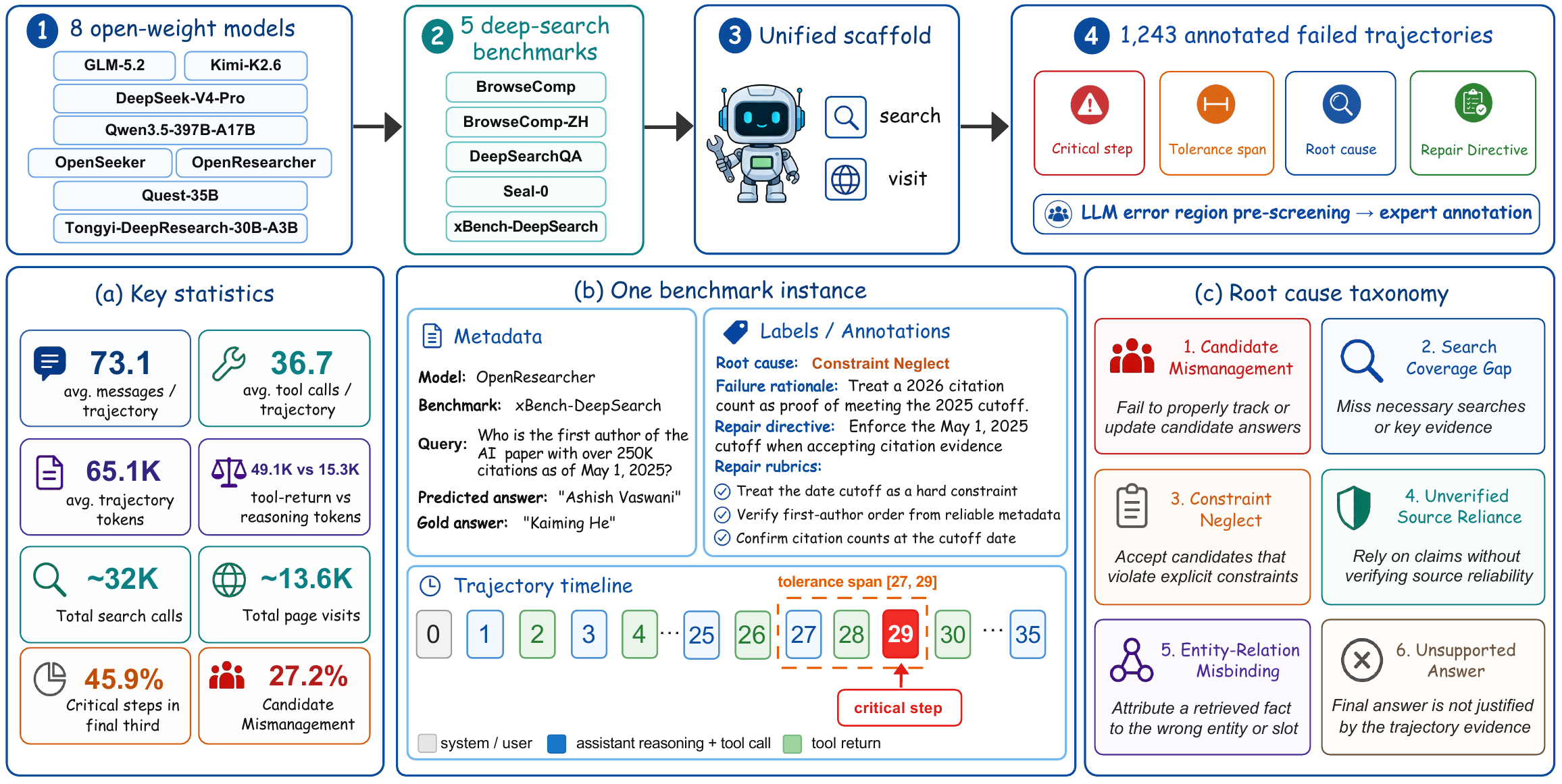}
    \caption{Overview of SearchAuditBench: construction pipeline (top), key statistics (a), one annotated instance (b), and the six-way root cause taxonomy (c).}
    \label{fig:dataset_figure}
\end{figure*}

\section{3. \ SearchAuditBench}

SearchAuditBench evaluates an LLM's ability to serve as a \emph{trajectory auditor} for search agents: given a failed search trajectory, the auditor must pinpoint \emph{where} the failure occurred, explain \emph{what} went wrong, and prescribe \emph{how} to repair it. Each instance is annotated with (i) a critical error step with a tolerance span, (ii) a root cause from a six-way failure taxonomy, and (iii) an expert-written repair directive paired with atomic grading rubrics.

\subsection{3.1 \ Trajectory Collection}


To construct SearchAuditBench, we restrict the model pool to open-weight models whose trajectories expose explicit intermediate reasoning, which is necessary for fine-grained trajectory auditing. Auditing tool-only or hidden-reasoning traces is an important but distinct setting, which we leave to future work. Under this criterion, we collect trajectories from eight models spanning two categories.
The first includes four large-scale general models: GLM-5.2~\cite{glm5team2026glm5vibecodingagentic}, Kimi-K2.6~\cite{kimiteam2026kimik25visualagentic}, DeepSeek-V4-Pro~\cite{deepseekai2026deepseekv4highlyefficientmilliontoken}, and Qwen3.5-397B-A17B~\cite{yang2025qwen3technicalreport}. The second includes four smaller models specialized for deep research: OpenSeeker~\cite{du2026openseeker}, OpenResearcher~\cite{li2026openresearcherfullyopenpipeline}, Quest-35B~\cite{xie2026questtrainingfrontierdeep}, and Tongyi-DeepResearch-30B-A3B~\cite{tongyideepresearchteam2026}. Queries are drawn from five deep-search benchmarks: BrowseComp~\cite{wei2025browsecompsimplechallengingbenchmark}, BrowseComp-ZH~\cite{zhou2025browsecompzhbenchmarkingwebbrowsing}, DeepSearchQA~\cite{gupta2026deepsearchqabridgingcomprehensivenessgap}, Seal-0~\cite{pham2026sealqaraisingbarreasoning}, and xBench-DeepSearch~\cite{chen2025xbenchtrackingagentsproductivity}. To isolate model behavior from scaffold effects, all eight models run under a unified scaffold adapted from Tongyi DeepResearch~\cite{tongyideepresearchteam2026}, which equips every model with two tools: \texttt{search} and \texttt{visit}. We retain only trace-auditable failures: trajectories whose final answers are judged as incorrect by an LLM-based evaluator against gold answers, and for which annotators can identify and justify the critical error using evidence contained in the frozen trajectory. This yields 1,243 failed trajectories, with a balanced per-model distribution.

\subsection{3.2 \  Task Design}
\label{sec:task}

Each benchmark input is a triple $x = (q, \hat{a}, \tau)$, where $q$ is the user query, $\tau = (m_1, \ldots, m_T)$ is the complete trajectory of interleaved reasoning and tool interactions, indexed by message position, and $\hat{a}$ is the incorrect final answer that $\tau$ terminates in, restated explicitly as the failure to be audited. We refer to the assistant messages of $\tau$ as its \emph{decision steps}: each one pairs a reasoning move with the tool call it issues, or emits the final answer $\hat{a}$. All gold and predicted step indices in this paper ($k^{*}$, $\hat{k}$) are message positions of decision steps; tool-return messages are never candidate steps. An auditor $\mathcal{A}$ must produce a structured diagnosis
\begin{equation}
\mathcal{A}(q, \hat{a}, \tau) \;=\; (\hat{k},\, \hat{c},\, \hat{r}),
\end{equation}
corresponding to three sub-tasks.


\paragraph{(1) Critical-step localization.}

The auditor must pinpoint where the failure originates by predicting a critical error step $\hat{k}$. We define the gold critical step $k^{*}$ as the earliest decision step at which the annotated failure mechanism becomes established, so that correcting $k^{*}$ would remove this mechanism and make a correct final answer attainable under competent subsequent search. Because the exact step at which the failure becomes established can be ambiguous, annotators also provide a tight tolerance span $[k_s, k_e]$: the smallest contiguous window of messages whose decision steps are all acceptable localizations of the critical error, with $k^{*} \in [k_s, k_e]$ by construction. Predictions are evaluated under both strict ($\hat{k} = k^{*}$) and loose ($\hat{k} \in [k_s, k_e]$) matching.


\paragraph{(2) Root-cause classification.}
We develop a six-way taxonomy of error categories (Figure~\ref{fig:dataset_figure}): Candidate Mismanagement, Search Coverage Gap, Constraint Neglect, Unverified Source Reliance, Entity-Relation Misbinding, and Unsupported Answer. The auditor must diagnose \emph{what} went wrong by assigning a root cause $\hat{c} \in \mathcal{C}$ from this taxonomy. The gold label $c^{*}$ is the expert-annotated root cause. Full definitions of the six categories are provided in Appendix A.

\paragraph{(3) Repair directive.}

The auditor must prescribe \emph{how} to repair the failure by producing $\hat{r}$, a concrete revision of the agent's behavior from the critical step onward. Repairs are graded against three to five expert-written atomic rubrics. Grading is \emph{diagnosis-gated}: $\hat{r}$ is scored only when $\hat{k}$ falls within the tolerance span and $\hat{c}$ is correct, so that repair quality is measured conditional on a sound diagnosis. Annotation details for all three tasks are given in Section~3.3.


\paragraph{Auditing setup.} 
Auditing is strictly offline: the auditor sees only $(q, \hat{a}, \tau)$ and knows the run failed, but receives neither the gold answer nor any web or tool access, so every claim must be grounded in the trajectory itself. We therefore evaluate only failures whose critical error is identifiable from the frozen trajectory (Section~3.1). This prevents diagnosis from collapsing into post-hoc problem solving, and keeps evaluation reproducible, since a frozen trajectory judges all auditors on identical evidence while live search results drift over time. Section 5 details the evaluation protocol and metrics.

\begin{figure*}[t]
    \centering
    \includegraphics[width=\textwidth]{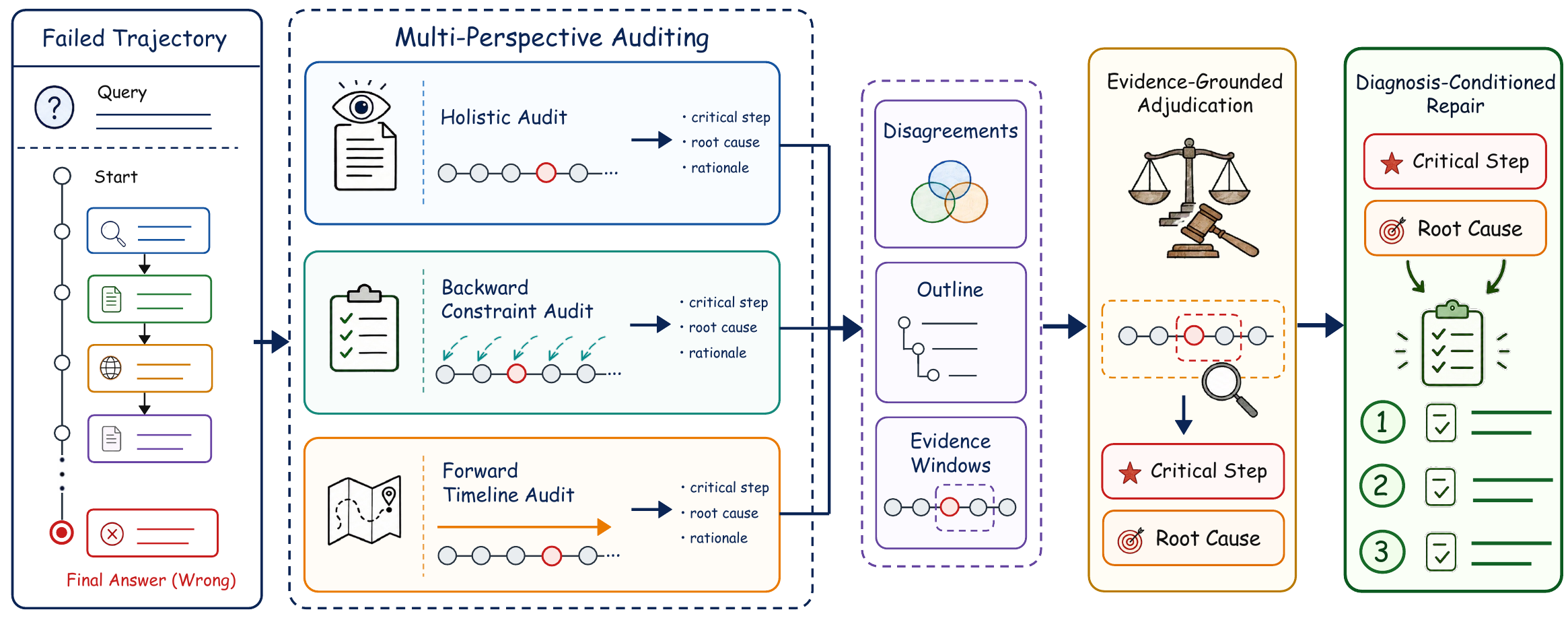}
    \caption{Overview of SearchAuditor: parallel multi-perspective auditing, evidence-grounded adjudication, and diagnosis-conditioned repair synthesis.}
    \label{fig:method_figure}
\end{figure*}



\subsection{3.3 \ Label Annotation}
\label{sec:annotation}

\paragraph{Annotation workflow.}

Four annotators from the author team, each with over one year of research experience on LLM agents, perform the annotation. To keep long trajectories tractable, an LLM assistant first proposes candidate error regions with rationales, which annotators verify, correct, or discard before finalizing labels. Every case is labeled by a single annotator, with cases split disjointly across the four annotators; more annotation details are in Appendix B.

\paragraph{Critical step and tolerance span.}
Annotators mark the critical step $k^{*}$ and its tolerance span $[k_s, k_e]$ following Section~3.2. When problematic evidence first appears in a tool return, they label the decision step that issued the call if the call itself was erroneous, otherwise the first decision step that misinterpreted or misused the returned evidence.



\paragraph{Root cause.}

When a trajectory contains multiple flaws, annotators label the single most direct, earliest failure-establishing error rather than its downstream symptoms, so each trajectory carries exactly one root cause.


\paragraph{Repair directives and rubrics.}
Annotators write a repair directive and decompose it into three to five atomic rubrics, each stating one requirement a valid repair must satisfy. Rubrics must be case-specific and process-level, constraining how the search is revised from the critical step onward rather than what the final answer is.

\subsection{3.4 \ Dataset Statistics}

\label{sec:statistics}

Figure 2(a) summarizes the key statistics of SearchAuditBench (extended statistics in the Appendix C). Failed trajectories are long and evidence-dominated: on average, each spans 73.1 messages and 65.1K tokens, of which 49.1K are tool returns against only 15.3K of captured reasoning, so the critical error is buried in raw evidence. Model behaviors also vary widely: per failed run, mean tool calls range from 7.4 (Tongyi-DeepResearch-30B-A3B) to 65.9 (GLM-5.2) and trajectory tokens from 24K to 118.3K, while the search-to-visit mix spans four searches per visit (Kimi-K2.6) to roughly one-to-one (OpenResearcher). Auditing is thus doubly challenging: an auditor must pinpoint a single critical step buried in long, noisy traces, while generalizing across sharply different trajectory shapes and tool-use patterns.




\section{4. \ SearchAuditor}
\label{sec:method}

Auditing a failed search trajectory requires pinpointing the earliest decision that established the failure within tens of thousands of tokens of accumulated evidence (Section 3.4). Such a failure is far more visible in its downstream symptoms, so a single pass over the trace tends to settle on a later step and leaves that verdict unchecked. SearchAuditor therefore decouples proposing candidate diagnoses from adjudicating them, mapping a failed trajectory to the complete audit output $(\hat{k}, \hat{c}, \hat{r})$ in three stages (Figure~\ref{fig:method_figure}). First, three audit branches analyze the same trajectory in parallel from complementary perspectives. Second, an adjudicator inspects the trajectory evidence around their proposed critical steps and jointly selects the critical step and root cause. Third, a repair synthesizer converts the fixed diagnosis into a small set of actionable, process-level directives.

\paragraph{Multi-perspective auditing.}
The three audit branches run in parallel and use the same model $f_{\theta}$, differing only in the audit procedure that each prompt $p_j$ enforces.
\begin{equation}
    z_j \;=\; f_{\theta}\!\left(q,\, \hat{a},\, \tau;\, p_j\right), \qquad j \in \{1, 2, 3\}.
\end{equation}
The use of distinct procedures is intended to expose perspective-dependent disagreements for the adjudicator to resolve. Unless otherwise stated, all stages share the same backbone: $f_{\theta}$ and $g_{\theta}$ denote role-specific prompting configurations of one model rather than separately trained components, with all prompts given in Appendix G.
Each report $z_j$ contains a candidate critical step, a root cause, and a failure rationale; the two specialized audits additionally include their intermediate analysis in their reports as audit notes for the adjudicator. The three procedures, enforced by $p_1$, $p_2$, and $p_3$ respectively, are as follows:
\begin{enumerate}[label=(\alph*), leftmargin=*, itemsep=1pt, topsep=2pt]
    \item The \emph{holistic audit} diagnoses the trajectory directly, without a prescribed decomposition, catching failures that the two specialized audits below might miss.
    \item The \emph{backward constraint audit} parses the query $q$ into a set of constraints 
    and traces each constraint unsupported by trajectory evidence back to the earliest decision where the agent committed to a search direction or candidate answer without first verifying it; when multiple constraints implicate different decisions, it reports the earliest one as its candidate critical step.
    \item The \emph{forward timeline audit} first derives a minimal query-grounded plan covering the explicit requirements of $q$, then walks through the decision steps in order and targets the earliest decision step that establishes the failure path rather than a later one that merely repeats or confirms the commitment.
\end{enumerate}

\paragraph{Evidence-grounded adjudication.}


Deterministic rules first organize the three reports into a structured adjudication context: disagreements are tabulated into a summary, and nearby critical-step proposals are grouped into clusters for comparison. Each distinct proposal only anchors where evidence is extracted and does not fix the final localization. Around every proposed critical step and the final-answer step, we extract a bounded window of original messages immediately surrounding the anchor. A single adjudicator $g_{\theta}$ then re-decides the two diagnosis fields, 
\begin{equation}
(\hat{k}, \hat{c}) = g_{\theta}\!\left(q,\,\hat{a},\,\tau,\,Z,\,O,\,W\right),
\end{equation}

where $Z$ collects the three audit reports together with the summary of their disagreements, and $O$ is a deterministic outline that lists every decision step's message position with a truncated content preview, supplying the global structure that the local windows $W$ lack. The adjudicator's decision must cite supporting messages and follow three rules: (i) agreement among reports is a diagnostic signal rather than a vote, and a minority report may prevail when the cited evidence supports it; (ii) the root cause and the critical step may be drawn from different reports, but the pair must be consistent, with the step being the earliest point at which the cause is instantiated; (iii) localization prefers the earliest supported decision step, and when an error surfaces in a tool output, it targets the decision step that issued an erroneous call, or otherwise the first decision step that misused the returned evidence, never the tool message itself.


\paragraph{Diagnosis-conditioned repair synthesis.}
A final synthesizer receives only $q$, $\hat{a}$, the adjudicated diagnosis with its rationale, and evidence windows newly extracted at the critical step and the final answer. Its output $\hat{r}$ consists of case-specific directives executable from the critical step onward, each describing a process change that addresses the diagnosed failure mechanism rather than stating the target answer.

\begin{table*}[ht]
\renewcommand{\arraystretch}{1.2}
\centering
\setlength{\tabcolsep}{4pt}
\small

\begin{tabular}{
  |>{\centering\arraybackslash}m{2.5cm}|
  >{\raggedright\arraybackslash}m{2.15cm}|
  cc|
  >{\centering\arraybackslash}m{1.25cm}
  >{\centering\arraybackslash}m{1.25cm}|
  >{\centering\arraybackslash}m{1.65cm}|
  >{\centering\arraybackslash}m{1.75cm}|
  >{\centering\arraybackslash}m{1.75cm}|
}
\hline

\multirow{2}{*}{\textbf{Model}}
  & \multirow{2}{*}{\textbf{Methods}}
  & \multicolumn{2}{c|}{\textbf{Critical Step}}
  & \multicolumn{2}{c|}{\textbf{Root Cause}}
  & \textbf{Diagnosis}
  & \textbf{Repair}
  & \textbf{End-to-End} \\

  &
  & \textit{CS-Strict}$\uparrow$
  & \textit{CS-Loose}$\uparrow$
  & \textit{RC-Acc}$\uparrow$
  & \textit{RC-F1}$\uparrow$
  & \textit{Diag}$\uparrow$
  & \textit{Rep@Diag}$\uparrow$
  & \textit{FPS}$\uparrow$ \\
\hline

\multirow{5}{*}{\textit{GPT-5.5}}
  & All-at-Once
  & 38.29
  & 51.89
  & 51.97
  & 49.06
  & 32.90
  & 80.68
  & 26.55 \\

  & Step-by-Step
  & 31.30
  & 45.53
  & 46.90
  & 42.42
  & 27.03
  & 79.46
  & 21.48 \\

  & Binary Search
  & 39.02
  & 51.33
  & 46.66
  & 44.63
  & 29.36
  & 83.29
  & 24.46 \\

  & AgentRx
  & 36.85
  & 51.01
  & 50.93
  & 48.51
  & 32.66
  & 54.19
  & 17.70 \\

\rowcolor{lightblue}
\cellcolor{white}
  & \textbf{SearchAuditor}
  & \textbf{44.89}
  & \textbf{58.73}
  & \textbf{54.79}
  & \textbf{51.10}
  & \textbf{38.05}
  & \textbf{84.78}
  & \textbf{32.26} \\
\hline

\multirow{5}{*}{\textit{Gemini-3.1-Pro}}
  & All-at-Once
  & 28.80
  & 43.77
  & 41.59
  & 36.75
  & 25.34
  & 53.65
  & 13.59 \\

  & Step-by-Step
  & 22.69
  & 35.96
  & 35.08
  & 30.19
  & 16.65
  & 58.53
  & 9.75 \\

  & Binary Search
  & 29.12
  & 41.19
  & 39.98
  & 36.45
  & 20.76
  & 55.81
  & 11.58 \\

  & AgentRx
  & 32.18
  & 45.05
  & 40.47
  & 36.17
  & 24.38
  & 42.24
  & 10.30 \\

\rowcolor{lightblue}
\cellcolor{white}
  & \textbf{SearchAuditor}
  & \textbf{36.12}
  & \textbf{48.43}
  & \textbf{44.17}
  & \textbf{39.06}
  & \textbf{30.57}
  & \textbf{61.84}
  & \textbf{18.91} \\
\hline

\multirow{5}{*}{\textit{Claude-Opus-4.8}}
  & All-at-Once
  & 32.10
  & 47.06
  & 47.63
  & 45.43
  & 28.24
  & 71.51
  & 20.19 \\

  & Step-by-Step
  & 23.09
  & 37.65
  & 37.49
  & 36.15
  & 18.91
  & 68.94
  & 13.03 \\

  & Binary Search
  & 32.82
  & 47.06
  & 43.36
  & 41.55
  & 25.34
  & 70.48
  & 17.86 \\

  & AgentRx
  & 35.40
  & 49.72
  & 42.96
  & 40.92
  & 26.71
  & 46.39
  & 12.39 \\

\rowcolor{lightblue}
\cellcolor{white}
  & \textbf{SearchAuditor}
  & \textbf{39.90}
  & \textbf{54.79}
  & \textbf{50.68}
  & \textbf{47.05}
  & \textbf{33.71}
  & \textbf{73.03}
  & \textbf{24.62} \\
\hline

\end{tabular}

\caption{
Main results on SearchAuditBench ($N=1{,}243$).
}
\label{tab:main}
\end{table*}
\section{5. \ Experiments}


\subsection{5.1 \ Experimental Setup}

\label{sec:exp_setup}

\paragraph{Baselines \& Models.} We compare SearchAuditor against two families of baselines. (i) \emph{Direct prompting}:  the three attribution strategies from \citet{zhang2025which}, each adapted to emit our full output triple (critical step, root cause, and repair directive). \textbf{All-at-Once} performs a single holistic judge call over the full trajectory; \textbf{Step-by-Step} walks the trajectory prefix by prefix and commits at the first critical error step it thinks decisive; and \textbf{Binary Search} recursively halves the trajectory to localize the critical segment. 
(ii) \emph{Framework}: \textbf{AgentRx} \cite{barke2026agentrxdiagnosingaiagent}, the closest prior framework to ours, which diagnoses a failed trajectory via an LLM judge over synthesized constraint violations; we adapt its constraint synthesis to the search agent's tool schema and extend its judge to emit our full output triple.
We instantiate every auditor on three frontier models: GPT-5.5 \cite{singh2026openaigpt5card}, Gemini-3.1-Pro \cite{googledeepmind2026gemini31pro}, and Claude-Opus-4.8 \cite{anthropic2026opus48}, with reasoning effort set to high. Repair rubrics are graded by DeepSeek-V4-Flash \cite{deepseekai2026deepseekv4highlyefficientmilliontoken}, which checks the predicted repair against each expert-written rubric and passes a case only if every rubric is satisfied; Appendix D validates this grader against human judgments, and Appendix E reports the cost and efficiency of all auditors.

\paragraph{Evaluation Metrics.}

All metrics are computed over the full benchmark ($N{=}1{,}243$) and reported in percentage points. For critical-step localization, CS-Strict is the fraction of cases whose predicted step exactly matches the gold critical step ($\hat{k} = k^{*}$), and CS-Loose relaxes this to falling within the annotated tolerance span ($\hat{k} \in [k_s, k_e]$). For root-cause classification, RC-Acc is exact-match accuracy over the six-way taxonomy and RC-F1 is the macro-average of per-class F1 scores over the six categories. Diag is the fraction of cases that achieve a \emph{sound diagnosis}, i.e., a correct root cause together with a predicted step inside the tolerance span. Rep@Diag is computed only over soundly diagnosed cases, as repair grading is diagnosis-gated: it is the fraction of those cases whose repair satisfies \emph{all} expert-written rubrics. Finally, the end-to-end fully-passed score FPS is the fraction of all cases that simultaneously achieve a sound diagnosis and a fully-passing repair, so that $\text{FPS} = \text{Diag} \times \text{Rep@Diag}$.


\subsection{5.2 \ Main Results}
\label{sec:main_results}

As shown in Table~\ref{tab:main}, we summarize key observations
below.

\paragraph{Universal performance improvement.} SearchAuditor outperforms every baseline on every metric under all three backbones. With GPT-5.5, it reaches 44.89\% CS-Strict and 58.73\% CS-Loose, exceeding the strongest baseline by 5.87 and 6.84 points, and raises the end-to-end FPS from 26.55\% to 32.26\%. The gains are not tied to a single model, holding at roughly 4 to 5 points in CS-Strict and FPS on the other two backbones (Table~\ref{tab:main}). They also cover both stages of the audit: SearchAuditor attains the best Diag and the best Rep@Diag in every block, improving Diag over the strongest baseline by more than five points on each backbone, so diagnosis and repair improvements multiply into the end-to-end score. These results highlight SearchAuditor's broad effectiveness across localization, attribution, repair, and its robustness to the choice of backbone.


\paragraph{Trajectory auditing remains challenging.} Even the best configuration, SearchAuditor with GPT-5.5, localizes the critical step exactly in fewer than half of the cases (44.89\% CS-Strict) and fully passes fewer than one third of audits (32.26\% FPS); even under the loose criterion, more than 40\% of cases are still mislocalized, and root-cause attribution stays around or below 55\% accuracy for all auditors. The baselines fall well below these levels: the strongest one reaches only 39.02\% CS-Strict and 26.55\% FPS, and with weaker backbones the end-to-end score drops to around 10\%. 
These results reflect the intrinsic difficulty of the task: as characterized in Section 3.4, failed trajectories are long, evidence-dominated, and nearly half of critical errors emerge only in the final third, demanding long-range integration before committing to a localization.

\subsection{5.3 \ Ablation Study}

\begin{table}[t]
\centering
\small
\setlength{\tabcolsep}{4pt}
\begin{tabular}{lcccc}
\toprule
Variant & CS-Strict$\uparrow$ & Diag$\uparrow$ & Rep@Diag$\uparrow$ & FPS$\uparrow$ \\
\midrule
Full (H+B+F) & \textbf{49.00} & \textbf{40.00} & \textbf{79.18} & \textbf{31.67} \\
\midrule
\emph{w/ 3$\times$Holistic} & 46.00 & 36.00 & 75.92 & 27.33 \\
\emph{w/o Forward} & 45.00 & 35.00 & 70.48 & 24.67 \\
\emph{w/o Backward} & 47.67 & 36.00 & 76.86 & 27.67 \\
\midrule
\emph{w/ Vote} & 44.33 & 34.00 & 74.51 & 25.33 \\
\emph{w/o $O$+$W$} & 44.67 & 37.00 & 72.97 & 27.00 \\
\bottomrule
\end{tabular}
\caption{Ablation of SearchAuditor on a 300-instance subset with GPT-5.5. H, B, F denote the holistic, backward constraint, and forward timeline audits.}
\label{tab:ablation}
\end{table}

We ablate SearchAuditor on a randomly sampled 300-instance subset with GPT-5.5 as the backbone (Table~\ref{tab:ablation}); all variants retain the three-stage pipeline. \emph{w/ 3$\times$Holistic} replaces both specialized audits with holistic copies (H+H+H) at identical compute, while \emph{w/o Forward} and \emph{w/o Backward} replace only one; \emph{w/ Vote} replaces the LLM adjudicator with deterministic aggregation (majority vote over root causes, fixed tie-breaking over nearby step proposals); \emph{w/o $O$+$W$} removes the outline and evidence windows from the adjudicator's input.

We have the following observations. (i) \textbf{Heterogeneous perspectives beat repeated sampling.} At identical compute, \emph{w/ 3$\times$Holistic} drops CS-Strict by 3.0 points and FPS by 4.3, so the gains come from complementary procedures rather than ensembling. Removing the forward timeline audit hurts most (4.0 points in CS-Strict, 7.0 in FPS), even more than removing both specialized audits, as it is the only branch that explicitly targets the step at which the failure becomes established. (ii) \textbf{Evidence-grounded adjudication outperforms voting.} \emph{w/ Vote} saves one model call but drops CS-Strict by 4.7 points and FPS by 6.3, indicating that cross-report disagreements must be resolved against trajectory evidence rather than by counting reports. (iii) \textbf{Structured evidence views matter even with the full trajectory.} \emph{w/o $O$+$W$} drops CS-Strict by 4.3 points and Rep@Diag by 6.2: the outline and windows act as structured anchors that concentrate adjudication on contested steps, which the full trajectory alone does not provide.

\subsection{5.4 \ Boosting Search Agents with Audits}
\label{sec:boost}
We next ask whether audits carry practical value: can they help a failing search agent recover? We inject audit outputs back into failed runs on a held-out benchmark, where the only signal is whether the resumed run reaches the correct answer, giving an annotation-free validation of diagnostic utility.

\paragraph{Setup.} We adopt LiveBrowseComp~\cite{fan2026livebrowsecompsearchagentssearching}, a deep-search benchmark of 335 questions whose answers depend on recently published facts, disjoint from the five source benchmarks of SearchAuditBench. We run Kimi-K2.6 and Quest-35B under the scaffold of Section 3.1. Their original runs solve 34.03\% and 8.96\% of the questions; the failed runs that terminate with a gradable final answer (213 and 300) form the repair cohorts, and the rest count as incorrect in Acc. Each failed trajectory is diagnosed with GPT-5.5 as the auditor backbone, and every intervention resumes the run by keeping the trajectory prefix up to the predicted critical step, appending the repair directive as a user message, and letting the agent continue under the original scaffold. We compare five interventions:

\begin{itemize}[leftmargin=*, itemsep=1pt, topsep=2pt]

    \item \emph{Unguided retry}: a fresh rerun from the original query, equivalent to pass@2 on the failed set.

    \item \emph{Generic hint}: a fixed prompt (``Reconsider your approach carefully before continuing.'') injected at SearchAuditor's predicted critical step, keeping its localization but stripping the diagnosis.

    \item \emph{Method repair}: the repair directive of All-at-Once, AgentRx, or SearchAuditor, injected at each method's own predicted critical step.

\end{itemize}

\begin{table}[ht]
\centering
\small
\setlength{\tabcolsep}{4pt}
\begin{tabular}{lcccc}
\toprule
& \multicolumn{2}{c}{Kimi-K2.6} & \multicolumn{2}{c}{Quest-35B} \\
\cmidrule(lr){2-3} \cmidrule(lr){4-5}
Intervention & Fix Rate$\uparrow$ & Acc.$\uparrow$ & Fix Rate$\uparrow$ & Acc.$\uparrow$ \\
\midrule
None (original run) &  & 34.03 &  & 8.96 \\
\midrule
Unguided retry & 9.39 & 40.00 & 5.00 & 13.43 \\
Generic hint & 5.63 & 37.61 & 4.67 & 13.13 \\
All-at-Once repair & 4.23 & 36.72 & 3.67 & 12.24 \\
AgentRx repair & 6.10 & 37.91 & 5.33 & 13.73 \\
SearchAuditor repair & \textbf{17.37} & \textbf{45.07} & \textbf{10.33} & \textbf{18.21} \\
\bottomrule
\end{tabular}
\caption{Repair-guided resumption on LiveBrowseComp. Fix Rate is over the 213 and 300 verifiable failed runs; Acc. is over all 335 questions.}
\label{tab:boost}
\end{table}

\paragraph{Recovery results.} SearchAuditor's repairs correct 17.37\% of Kimi-K2.6 and 10.33\% of Quest-35B failures, roughly doubling unguided retry and lifting overall accuracy to 45.07\% and 18.21\% (Table~\ref{tab:boost}). No other guided intervention clearly improves on unguided retry, which we attribute to resumption asymmetry: a resumed run inherits the prefix that produced the failure, so inaccurate guidance anchors the agent to its flawed reasoning, whereas a fresh restart at least escapes it. Notably, the generic hint shares SearchAuditor's localization exactly yet trails it by 11.74 and 5.66 points, confirming that the repair, not merely the localization, drives the recovery gain.

\subsection{5.5\ Analysis of Search-Agent Failures}

\label{sec:failure_anatomy}
 
Beyond enabling auditor evaluation, the corpus reveals how long-horizon
search agents fail. We highlight three findings.

\paragraph{Most failures are not retrieval failures.}
Only 22.8\% of failures stem from \emph{Search Coverage Gap}, where the decisive evidence was never retrieved. In the remaining 77.2\%, the primary error lies not in what the agent retrieved but in how it processed what it had, led by \emph{Candidate Mismanagement} (27.2\% of the corpus) and \emph{Constraint Neglect} (19.1\%). Indeed, in 25.0\% of all failures the correct answer appears verbatim in the retrieved content, and in 83\% of these it was retrieved at or before the critical step. This ``answer-in-hand'' rate reaches 35.6\% for \emph{Candidate Mismanagement} but only 12.4\% for \emph{Search Coverage Gap}, indicating that the primary bottleneck is evidence utilization rather than retrieval.
 

\paragraph{Failures propagate silently and waste nearly half of agent
compute.} Search agents rarely halt after going wrong. After the critical error step, they continue for a mean of 16.7 further tool calls. Corpus-wide, 47.1\% of generated tokens and 45.4\% of tool calls occur \emph{after} the decisive error, without ever repairing it. Reliable mid-trajectory auditing could in principle reclaim this wasted compute, consistent with the recovery gains from repair-guided resumption in Section~5.4.

\begin{figure}[h]
    \centering
    \includegraphics[width=1\linewidth]{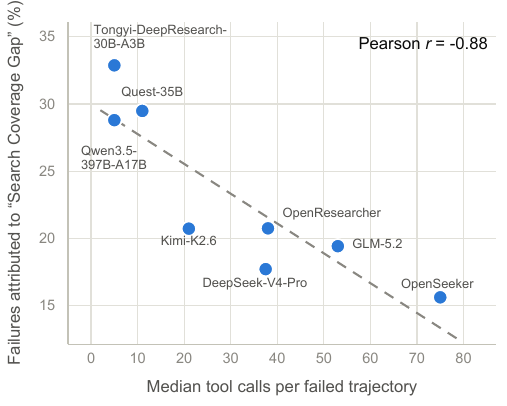}
    \caption{Search Coverage Gap failures (\%) vs.\ median tool calls per failed trajectory across the eight agents.}
    \label{fig:budget_shift}
\end{figure}

\paragraph{Scaling exploration moves failures downstream rather than
removing them.}
Exploration budgets vary widely across agents, from a median of 5 tool calls per failed run to 75. As Figure~\ref{fig:budget_shift} shows, the fraction of failures attributed to \emph{Search Coverage Gap} correlates strongly and negatively with this budget (Pearson $r = -0.88$): sparse explorers fail by not looking (29\% to 33\% \emph{Search Coverage Gap}), while heavy explorers largely eliminate coverage failures (16\% to 19\%) but shift toward \emph{Candidate Mismanagement} and \emph{Constraint Neglect}, which together exceed half of their failures. Test-time scaling of search is therefore self-limiting: beyond a certain budget, the binding constraint shifts from evidence coverage to the agent's ability to manage candidates and constraints over an ever-longer context.
\section{6. \ Conclusion}

In this work, we presented a systematic study of auditing and attributing failures in long-horizon search agents. We constructed SearchAuditBench, an expert-annotated benchmark of 1,243 failed trajectories labeled with critical error steps, root causes, and repair rubrics, and proposed SearchAuditor, a multi-perspective auditing framework with evidence-grounded adjudication. Extensive experiments show that SearchAuditor consistently outperforms all baselines across three frontier backbones, and that resuming failed runs with its repairs enables agents to effectively recover from errors. Beyond evaluating auditors, the corpus also offers an empirical anatomy of how long-horizon search agents fail.

\bibliography{aaai2027}

@misc{wei2025browsecompsimplechallengingbenchmark,
      title={BrowseComp: A Simple Yet Challenging Benchmark for Browsing Agents}, 
      author={Jason Wei and Zhiqing Sun and Spencer Papay and Scott McKinney and Jeffrey Han and Isa Fulford and Hyung Won Chung and Alex Tachard Passos and William Fedus and Amelia Glaese},
      year={2025},
      eprint={2504.12516},
      archivePrefix={arXiv},
      primaryClass={cs.CL},
      url={https://arxiv.org/abs/2504.12516}, 
}

@article{nakano2021webgpt,
  title={Webgpt: Browser-assisted question-answering with human feedback},
  author={Nakano, Reiichiro and Hilton, Jacob and Balaji, Suchir and Wu, Jeff and Ouyang, Long and Kim, Christina and Hesse, Christopher and Jain, Shantanu and Kosaraju, Vineet and Saunders, William and others},
  journal={arXiv preprint arXiv:2112.09332},
  year={2021}
}

@misc{xie2026questtrainingfrontierdeep,
      title={QUEST: Training Frontier Deep Research Agents with Fully Synthetic Tasks}, 
      author={Jian Xie and Tianhe Lin and Zilu Wang and Yuting Ning and Yuekun Yao and Tianci Xue and Zhehao Zhang and Zhongyang Li and Kai Zhang and Yufan Wu and Shijie Chen and Boyu Gou and Mingzhe Han and Yifei Wang and Vint Lee and Xinpeng Wei and Xiangjun Wang and Yu Su and Huan Sun},
      year={2026},
      eprint={2605.24218},
      archivePrefix={arXiv},
      primaryClass={cs.CL},
      url={https://arxiv.org/abs/2605.24218}, 
}

@misc{li2026openresearcherfullyopenpipeline,
      title={OpenResearcher: A Fully Open Pipeline for Long-Horizon Deep Research Trajectory Synthesis}, 
      author={Zhuofeng Li and Dongfu Jiang and Xueguang Ma and Haoxiang Zhang and Ping Nie and Yuyu Zhang and Kai Zou and Jianwen Xie and Yu Zhang and Wenhu Chen},
      year={2026},
      eprint={2603.20278},
      archivePrefix={arXiv},
      primaryClass={cs.IR},
      url={https://arxiv.org/abs/2603.20278}, 
}

@misc{glm5team2026glm5vibecodingagentic,
      title={GLM-5: from Vibe Coding to Agentic Engineering}, 
      author={GLM-5-Team},
      year={2026},
      eprint={2602.15763},
      archivePrefix={arXiv},
      primaryClass={cs.LG},
      url={https://arxiv.org/abs/2602.15763}, 
}

@misc{zhao2026failureprocessanatomycli,
      title={Failure as a Process: An Anatomy of CLI Coding Agent Trajectories}, 
      author={Xiangxin Zhao and Han Li and Shuaiting Li and Tianyi Zhao and Earl T. Barr and Federica Sarro and He Ye},
      year={2026},
      eprint={2607.09510},
      archivePrefix={arXiv},
      primaryClass={cs.SE},
      url={https://arxiv.org/abs/2607.09510}, 
}

@misc{deshpande2025trailtracereasoningagentic,
      title={TRAIL: Trace Reasoning and Agentic Issue Localization}, 
      author={Darshan Deshpande and Varun Gangal and Hersh Mehta and Jitin Krishnan and Anand Kannappan and Rebecca Qian},
      year={2025},
      eprint={2505.08638},
      archivePrefix={arXiv},
      primaryClass={cs.AI},
      url={https://arxiv.org/abs/2505.08638}, 
}

@misc{fan2026livebrowsecompsearchagentssearching,
      title={LiveBrowseComp: Are Search Agents Searching, or Just Verifying What They Already Know?}, 
      author={HuiMing Fan and Xiao Wang and Zheng Chu and Qianyu Wang and Zhuoyao Wang and Ming Liu and Bing Qin and XingYu},
      year={2026},
      eprint={2605.28721},
      archivePrefix={arXiv},
      primaryClass={cs.AI},
      url={https://arxiv.org/abs/2605.28721}, 
}

@misc{googledeepmind2026gemini31pro,
  author  = {{Google DeepMind}},
  title   = {Gemini 3.1 Pro: Model Card},
  year    = {2026},
  url     = {https://deepmind.google/models/model-cards/gemini-3-1-pro/},
  urldate = {2026-07-19}
}

@misc{anthropic2026opus48,
  title={Claude Opus 4.8 System Card},
  author={Anthropic},
  year={2026},
  url={https://www-cdn.anthropic.com/0b4915911bb0d19eca5b5ee635c80fef830a37ea.pdf}
}

@misc{singh2026openaigpt5card,
      title={OpenAI GPT-5 System Card}, 
      author={OpenAI},
      year={2026},
      eprint={2601.03267},
      archivePrefix={arXiv},
      primaryClass={cs.CL},
      url={https://arxiv.org/abs/2601.03267}, 
}

@article{li2025websailor,
  title={Websailor: Navigating super-human reasoning for web agent},
  author={Li, Kuan and Zhang, Zhongwang and Yin, Huifeng and Zhang, Liwen and Ou, Litu and Wu, Jialong and Yin, Wenbiao and Li, Baixuan and Tao, Zhengwei and Wang, Xinyu and others},
  journal={arXiv preprint arXiv:2507.02592},
  year={2025}
}

@misc{zhan2026deepresearchagentfails,
      title={Why Your Deep Research Agent Fails? On Hallucination Evaluation in Full Research Trajectory}, 
      author={Yuhao Zhan and Tianyu Fan and Linxuan Huang and Zirui Guo and Chao Huang},
      year={2026},
      eprint={2601.22984},
      archivePrefix={arXiv},
      primaryClass={cs.AI},
      url={https://arxiv.org/abs/2601.22984}, 
}

@inproceedings{
zhang2025which,
title={Which Agent Causes Task Failures and When? On Automated Failure Attribution of {LLM} Multi-Agent Systems},
author={Shaokun Zhang and Ming Yin and Jieyu Zhang and Jiale Liu and Zhiguang Han and Jingyang Zhang and Beibin Li and Chi Wang and Huazheng Wang and Yiran Chen and Qingyun Wu},
booktitle={Forty-second International Conference on Machine Learning},
year={2025},
url={https://openreview.net/forum?id=GazlTYxZss}
}

@misc{zhang2025agentracerinducingfailurellm,
      title={AgenTracer: Who Is Inducing Failure in the LLM Agentic Systems?}, 
      author={Guibin Zhang and Junhao Wang and Junjie Chen and Wangchunshu Zhou and Kun Wang and Shuicheng Yan},
      year={2025},
      eprint={2509.03312},
      archivePrefix={arXiv},
      primaryClass={cs.CL},
      url={https://arxiv.org/abs/2509.03312}, 
}

@inproceedings{
cemri2026why,
title={Why Do Multi-Agent {LLM} Systems Fail?},
author={Mert Cemri and Melissa Z Pan and Shuyi Yang and Lakshya A Agrawal and Bhavya Chopra and Rishabh Tiwari and Kurt Keutzer and Aditya Parameswaran and Dan Klein and Kannan Ramchandran and Matei Zaharia and Joseph E. Gonzalez and Ion Stoica},
booktitle={The Thirty-ninth Annual Conference on Neural Information Processing Systems Datasets and Benchmarks Track},
year={2026},
url={https://openreview.net/forum?id=fAjbYBmonr}
}

@article{gou2026mind2web,
  title={Mind2web 2: Evaluating agentic search with agent-as-a-judge},
  author={Gou, Boyu and Huang, Zanming and Ning, Yuting and Gu, Yu and Lin, Michael and Qi, Weijian and Kopanev, Andrei and Yu, Botao and Jimenez Gutierrez, Bernal and Shu, Yiheng and others},
  journal={Advances in Neural Information Processing Systems},
  volume={38},
  year={2026}
}

@inproceedings{krishna-etal-2025-fact,
    title = "Fact, Fetch, and Reason: A Unified Evaluation of Retrieval-Augmented Generation",
    author = "Krishna, Satyapriya  and
      Krishna, Kalpesh  and
      Mohananey, Anhad  and
      Schwarcz, Steven  and
      Stambler, Adam  and
      Upadhyay, Shyam  and
      Faruqui, Manaal",
    editor = "Chiruzzo, Luis  and
      Ritter, Alan  and
      Wang, Lu",
    booktitle = "Proceedings of the 2025 Conference of the Nations of the Americas Chapter of the Association for Computational Linguistics: Human Language Technologies (Volume 1: Long Papers)",
    month = apr,
    year = "2025",
    address = "Albuquerque, New Mexico",
    publisher = "Association for Computational Linguistics",
    url = "https://aclanthology.org/2025.naacl-long.243/",
    doi = "10.18653/v1/2025.naacl-long.243",
    pages = "4745--4759",
    ISBN = "979-8-89176-189-6"
}

@inproceedings{chen-etal-2026-beyond-single,
    title = "Beyond Single-shot Writing: Deep Research Agents are Unreliable at Multi-turn Report Revision",
    author = "Chen, Bingsen  and
      Li, Boyan  and
      Nie, Ping  and
      Zhang, Yuyu  and
      Ye, Xi  and
      Zhao, Chen",
    editor = "Liakata, Maria  and
      Moreira, Viviane P.  and
      Zhang, Jiajun  and
      Jurgens, David",
    booktitle = "Proceedings of the 64th Annual Meeting of the {A}ssociation for {C}omputational {L}inguistics (Volume 1: Long Papers)",
    month = jul,
    year = "2026",
    address = "San Diego, California, United States",
    publisher = "Association for Computational Linguistics",
    url = "https://aclanthology.org/2026.acl-long.609/",
    doi = "10.18653/v1/2026.acl-long.609",
    pages = "13325--13356",
    ISBN = "979-8-89176-390-6"
}

@article{rafi2026falat,
  title={FALAT: Tracing Failures in LLM Agent Trajectories via Dependency-Guided Search},
  author={Rafi, Md Nakhla and Ahasanuzzaman, Md and Kim, Dong Jae and Wang, Zhijie and Chen, Tse-Hsun},
  journal={arXiv preprint arXiv:2606.00765},
  year={2026}
}

@misc{wang2026trajauditautomatedfailurediagnosis,
      title={TrajAudit: Automated Failure Diagnosis for Agentic Coding Systems}, 
      author={Minxing Wang and Xiaofei Xie and Yintong Huo},
      year={2026},
      eprint={2605.26563},
      archivePrefix={arXiv},
      primaryClass={cs.SE},
      url={https://arxiv.org/abs/2605.26563}, 
}

@misc{li2026codetracertraceableagentstates,
      title={CodeTracer: Towards Traceable Agent States}, 
      author={Han Li and Yifan Yao and Letian Zhu and Rili Feng and Hongyi Ye and Jiaming Wang and Yancheng He and Pengyu Zou and Lehan Zhang and Xinping Lei and Haoyang Huang and Ken Deng and Ming Sun and Zhaoxiang Zhang and He Ye and Jiaheng Liu},
      year={2026},
      eprint={2604.11641},
      archivePrefix={arXiv},
      primaryClass={cs.SE},
      url={https://arxiv.org/abs/2604.11641}, 
}

@article{deng2026memtrace,
  title={MemTrace: Tracing and Attributing Errors in Large Language Model Memory Systems},
  author={Deng, Xinle and Zhong, Ruobin and Peng, Hujin and Lu, Xiaoben and Wu, Yanzhe and Li, Guang and Xu, Buqiang and Yao, Yunzhi and Fang, Jizhan and Cao, Haoliang and others},
  journal={arXiv preprint arXiv:2605.28732},
  year={2026}
}

@misc{kimiteam2026kimik25visualagentic,
      title={Kimi K2.5: Visual Agentic Intelligence}, 
      author={Kimi-Team},
      year={2026},
      eprint={2602.02276},
      archivePrefix={arXiv},
      primaryClass={cs.CL},
      url={https://arxiv.org/abs/2602.02276}, 
}

@misc{barke2026agentrxdiagnosingaiagent,
      title={AgentRx: Diagnosing AI Agent Failures from Execution Trajectories}, 
      author={Shraddha Barke and Arnav Goyal and Alind Khare and Avaljot Singh and Suman Nath and Chetan Bansal},
      year={2026},
      eprint={2602.02475},
      archivePrefix={arXiv},
      primaryClass={cs.AI},
      url={https://arxiv.org/abs/2602.02475}, 
}

@misc{deepseekai2026deepseekv4highlyefficientmilliontoken,
      title={DeepSeek-V4: Towards Highly Efficient Million-Token Context Intelligence}, 
      author={DeepSeek-AI},
      year={2026},
      eprint={2606.19348},
      archivePrefix={arXiv},
      primaryClass={cs.CL},
      url={https://arxiv.org/abs/2606.19348}, 
}

@article{du2026openseeker,
  title={Openseeker: Democratizing frontier search agents by fully open-sourcing training data},
  author={Du, Yuwen and Ye, Rui and Tang, Shuo and Zhu, Xinyu and Lu, Yijun and Cai, Yuzhu and Chen, Siheng},
  journal={arXiv preprint arXiv:2603.15594},
  year={2026}
}

@misc{yang2025qwen3technicalreport,
      title={Qwen3 Technical Report}, 
      author={An Yang and Anfeng Li and Baosong Yang and Beichen Zhang and Binyuan Hui and Bo Zheng and Bowen Yu and Chang Gao and Chengen Huang and Chenxu Lv and Chujie Zheng and Dayiheng Liu and Fan Zhou and Fei Huang and Feng Hu and Hao Ge and Haoran Wei and Huan Lin and Jialong Tang and Jian Yang and Jianhong Tu and Jianwei Zhang and Jianxin Yang and Jiaxi Yang and Jing Zhou and Jingren Zhou and Junyang Lin and Kai Dang and Keqin Bao and Kexin Yang and Le Yu and Lianghao Deng and Mei Li and Mingfeng Xue and Mingze Li and Pei Zhang and Peng Wang and Qin Zhu and Rui Men and Ruize Gao and Shixuan Liu and Shuang Luo and Tianhao Li and Tianyi Tang and Wenbiao Yin and Xingzhang Ren and Xinyu Wang and Xinyu Zhang and Xuancheng Ren and Yang Fan and Yang Su and Yichang Zhang and Yinger Zhang and Yu Wan and Yuqiong Liu and Zekun Wang and Zeyu Cui and Zhenru Zhang and Zhipeng Zhou and Zihan Qiu},
      year={2025},
      eprint={2505.09388},
      archivePrefix={arXiv},
      primaryClass={cs.CL},
      url={https://arxiv.org/abs/2505.09388}, 
}

@misc{tongyideepresearchteam2026,
      title={Tongyi DeepResearch Technical Report}, 
      author={Tongyi-DeepResearch-Team and Baixuan Li and Bo Zhang and Dingchu Zhang and Fei Huang and Guangyu Li and Guoxin Chen and Huifeng Yin and Jialong Wu and Jingren Zhou and Kuan Li and Liangcai Su and Litu Ou and Liwen Zhang and Pengjun Xie and Rui Ye and Wenbiao Yin and Xinmiao Yu and Xinyu Wang and Xixi Wu and Xuanzhong Chen and Yida Zhao and Zhen Zhang and Zhengwei Tao and Zhongwang Zhang and Zile Qiao and Chenxi Wang and Donglei Yu and Gang Fu and Haiyang Shen and Jiayin Yang and Jun Lin and Junkai Zhang and Kui Zeng and Li Yang and Hailong Yin and Maojia Song and Ming Yan and Minpeng Liao and Peng Xia and Qian Xiao and Rui Min and Ruixue Ding and Runnan Fang and Shaowei Chen and Shen Huang and Shihang Wang and Shihao Cai and Weizhou Shen and Xiaobin Wang and Xin Guan and Xinyu Geng and Yingcheng Shi and Yuning Wu and Zhuo Chen and Zijian Li and Yong Jiang},
      year={2026},
      eprint={2510.24701},
      archivePrefix={arXiv},
      primaryClass={cs.CL},
      url={https://arxiv.org/abs/2510.24701}, 
}

@misc{zhou2025browsecompzhbenchmarkingwebbrowsing,
      title={BrowseComp-ZH: Benchmarking Web Browsing Ability of Large Language Models in Chinese}, 
      author={Peilin Zhou and Bruce Leon and Xiang Ying and Can Zhang and Yifan Shao and Qichen Ye and Dading Chong and Zhiling Jin and Chenxuan Xie and Meng Cao and Yuxin Gu and Sixin Hong and Jing Ren and Jian Chen and Chao Liu and Yining Hua},
      year={2025},
      eprint={2504.19314},
      archivePrefix={arXiv},
      primaryClass={cs.CL},
      url={https://arxiv.org/abs/2504.19314}, 
}

@misc{pham2026sealqaraisingbarreasoning,
      title={SealQA: Raising the Bar for Reasoning in Search-Augmented Language Models}, 
      author={Thinh Pham and Nguyen Nguyen and Pratibha Zunjare and Weiyuan Chen and Yu-Min Tseng and Tu Vu},
      year={2026},
      eprint={2506.01062},
      archivePrefix={arXiv},
      primaryClass={cs.CL},
      url={https://arxiv.org/abs/2506.01062}, 
}

@misc{gupta2026deepsearchqabridgingcomprehensivenessgap,
      title={DeepSearchQA: Bridging the Comprehensiveness Gap for Deep Research Agents}, 
      author={Nikita Gupta and Riju Chatterjee and Lukas Haas and Connie Tao and Andrew Wang and Chang Liu and Hidekazu Oiwa and Elena Gribovskaya and Jan Ackermann and John Blitzer and Sasha Goldshtein and Dipanjan Das},
      year={2026},
      eprint={2601.20975},
      archivePrefix={arXiv},
      primaryClass={cs.CL},
      url={https://arxiv.org/abs/2601.20975}, 
}

@misc{chen2025xbenchtrackingagentsproductivity,
      title={xbench: Tracking Agents Productivity Scaling with Profession-Aligned Real-World Evaluations}, 
      author={Kaiyuan Chen and Yixin Ren and Yang Liu and Xiaobo Hu and Haotong Tian and Tianbao Xie and Fangfu Liu and Haoye Zhang and Hongzhang Liu and Yuan Gong and Chen Sun and Han Hou and Hui Yang and James Pan and Jianan Lou and Jiayi Mao and Jizheng Liu and Jinpeng Li and Kangyi Liu and Kenkun Liu and Rui Wang and Run Li and Tong Niu and Wenlong Zhang and Wenqi Yan and Xuanzheng Wang and Yuchen Zhang and Yi-Hsin Hung and Yuan Jiang and Zexuan Liu and Zihan Yin and Zijian Ma and Zhiwen Mo},
      year={2025},
      eprint={2506.13651},
      archivePrefix={arXiv},
      primaryClass={cs.LG},
      url={https://arxiv.org/abs/2506.13651}, 
}

\clearpage
\onecolumn
\lstset{numbers=none,xleftmargin=0pt}
\appendix

\newcommand{\appsection}[1]{%
    \section[#1]{\raggedright #1}%
}

\section*{\huge Appendix}

This appendix is organized as follows.

\begin{itemize}
    \item In Section A, we present the root-cause taxonomy development and definitions.
    \item In Section B, we detail the annotation pipeline, including the annotation guideline and interface.
    \item In Section C, we report extended statistics of SearchAuditBench.
    \item In Section D, we validate the automatic repair-rubric grader against a blind human reference.
    \item In Section E, we analyze the inference cost and latency of every auditor on a fixed subset.
    \item In Section F, we present two case studies.
    \item In Section G, we provide the complete prompts of SearchAuditor.
\end{itemize}

\appsection{A. \ Root-Cause Taxonomy Development and Definitions}
\newcommand{\rc}[1]{\emph{#1}}

\paragraph{Development procedure.}
We derived the taxonomy bottom-up from observed failures rather than imposing a stage-based scheme in advance. Pilot annotation with a deliberately fine-grained codebook exposed two problems as annotation scaled: some labels became catch-alls, since a wrong candidate could reflect premature commitment on partial matches, acceptance of a constraint violation, or rationalization of contradictory evidence; others described different manifestations of one repairable mechanism, producing overlap and sparse tail classes. We therefore consolidated premature commitment, erroneous rejection, failure to update, and failure to converge under \rc{Candidate Mismanagement}, and acceptance of and rationalization about hard-constraint violations under \rc{Constraint Neglect}, while retaining the already separable \rc{Search Coverage Gap}, \rc{Unverified Source Reliance}, \rc{Entity--Relation Misbinding}, and the residual \rc{Unsupported Answer}. Consolidation was followed by trajectory-level re-adjudication rather than a deterministic remapping, and the residual disagreements sharpened the boundaries stated below.

\paragraph{Labeling policy.}
Each trajectory receives exactly one primary root cause: the earliest and most direct mechanism that establishes the failure path, so downstream symptoms never override an earlier causal error. Operational behaviors such as ineffective queries, premature stopping, or extensive but unproductive search are manifestations of a mechanism rather than causes in themselves, and the stage at which an error occurs is kept separate from the mechanism explaining why it occurs. 

\paragraph{Category definitions.}
The six categories are defined below. Each definition opens with the criterion applied at annotation time, continues with the typical cases observed in the corpus, and closes with the boundary against the categories it is most easily confused with.

\begin{itemize}
\item \textbf{\rc{Candidate Mismanagement}.} The agent fails to maintain, compare, or update candidate answers correctly after candidates or candidate-relevant evidence have entered the trajectory. Typical cases include prematurely locking onto a candidate that matches only a salient subset of the clues, discarding a still-viable candidate without adequate comparison, failing to compare multiple candidates systematically, failing to update the answer after stronger evidence appears, and searching extensively without converting accumulated evidence into a candidate decision. The defining failure concerns the state and comparison of the candidate set; if the decisive candidate or evidence path never appears because it was never searched, the label is \rc{Search Coverage Gap} instead.

\item \textbf{\rc{Search Coverage Gap}.} The agent does not search for or inspect a key evidence path needed to solve the problem. Typical cases include an overly narrow search direction, failure to explore obvious candidates, failure to inspect a key data source, omission of a key evidence table or official source, never seeking the evidence that a decisive query constraint requires, and terminating before the relevant portion of the search space is examined. The defining property is the absence of the necessary evidence from the trajectory; if the source or candidate was found but subsequently mishandled, a source- or candidate-related category applies instead.

\item \textbf{\rc{Constraint Neglect}.} The agent accepts a candidate that does not satisfy a hard query constraint. Typical cases include ignoring a known contradiction, relaxing an explicit condition, treating a mismatch as acceptable, and rationalizing contradictory evidence so that the preferred candidate can be retained. A failed constraint alone is not sufficient for this label: if the information needed to evaluate the constraint was never retrieved, the cause is normally \rc{Search Coverage Gap}; if that information was attached to the wrong entity or relation, it is \rc{Entity--Relation Misbinding}.

\item \textbf{\rc{Unverified Source Reliance}.} The agent finds an apparently relevant source but does not verify source reliability or original content before relying on it. Typical cases include trusting snippets or titles as conclusive, relying on a secondary or low-quality source where primary evidence is required, accepting a claim from an inaccessible page, and committing to a candidate without opening the underlying primary source. The defining failure is insufficient verification of a source that was found, rather than the absence of a relevant search path.

\item \textbf{\rc{Entity--Relation Misbinding}.} The agent finds relevant facts but binds them to the wrong entity, relation, role, time, work, person, organization, or answer slot. Typical cases include conflating same-name entities, transferring an attribute from one person or organization to another, reversing a relation, and combining facts drawn from different candidate chains. The core problem is that the fact, possibly correct in itself, was attached to the wrong target rather than that no fact was found: the needed information is present, which separates this category from \rc{Search Coverage Gap}, and source reliability is not the central problem, which separates it from \rc{Unverified Source Reliance}.

\item \textbf{\rc{Unsupported Answer}.} The final answer lacks support from the trajectory evidence, and no more specific search coverage, source verification, candidate management, constraint neglect, or entity--relation binding failure explains it. Typical cases include guessing, memory-based completion, asserting a value not extracted from any inspected source, and extracting an answer from partial observations without an evidence chain. It is a residual category rather than a default label, assigned only after the other five mechanisms have been ruled out.
\end{itemize}

\paragraph{Boundary summary.}
Applied at the critical step, the categories are separated by the state of the decisive evidence. If it was never sought or inspected, the failure is a \rc{Search Coverage Gap}; if it was found but not verified, \rc{Unverified Source Reliance}; if candidates were surfaced but improperly maintained or compared, \rc{Candidate Mismanagement}; if an explicit hard-constraint mismatch was available yet accepted or rationalized, \rc{Constraint Neglect}; if a relevant fact was attached to the wrong target, \rc{Entity--Relation Misbinding}. Only when none of these earlier mechanisms accounts for the failure do we assign \rc{Unsupported Answer}.

\appsection{B. Annotation Details}

\label{app:annotation}

\subsection{B.1 \ Trajectory Pool and Discard Criteria}

\paragraph{Collection and automatic screening.} We draw a fixed query pool from each of the five source benchmarks and run all eight agents once on every sampled query, so that all 40 model $\times$ benchmark combinations contribute a comparable number of rollouts; the counts are only approximately equal because some released query sets are smaller than the target pool size. This yields 3,500 raw trajectories, of which 3,288 (93.9\%) terminate normally with a gradable final answer, while the remainder exhaust the context budget, hit the maximum turn limit, or abort on an unrecoverable tool error. An LLM evaluator then compares every gradable answer against the benchmark gold answer, marking 1,674 runs (50.9\% of gradable runs) as incorrect. These 1,674 failed runs form the pool that enters expert annotation.

\paragraph{Discard criteria.} Not every incorrect run is auditable from its frozen trace, so annotators discard a case whenever one of the following conditions holds, recording the condition rather than producing labels: \textbf{D1 (evaluator false negative)}, where the predicted answer is in fact acceptable and differs from the gold answer only by an alias, a unit, a rounding convention, or a formatting choice; \textbf{D2 (flawed benchmark instance)}, where the query is genuinely ambiguous, admits several defensible answers, or its gold answer is stale relative to the evidence the agent could observe; \textbf{D3 (incomplete trajectory)}, where the recorded trace is corrupted or partially missing, most often through mid-run context truncation or absent \texttt{reasoning\_content}; \textbf{D4 (not offline-auditable)}, where the decisive evidence never enters the trace and the failure is attributable to the environment rather than to an agent decision, for example when the search backend returns empty results throughout or every candidate page is unreachable; and \textbf{D5 (no single critical step)}, where several independent errors of comparable causal weight coexist and no single decision can be identified as the earliest one that establishes the failure path, so the single-primary-cause labeling policy of Appendix A cannot be applied. Discarding these cases from the pooled failed runs leaves the 1,243 fully annotated trajectories of SearchAuditBench, with 125 to 175 trajectories per generating agent.

\subsection{B.2 \ Annotation Guideline and Interface}

\paragraph{Annotation workflow.} Every case is labeled by a single annotator, with cases split disjointly across the four annotators. Because a failed trajectory averages 73.1 messages and 65.1K tokens, reading each trace end to end without guidance is prohibitively slow, so we precede human annotation with an LLM pre-screening pass. Claude Opus 4.6 receives the query, the predicted answer, the gold answer, and the complete trajectory, and proposes a small set of candidate error regions, each with a short rationale and the taxonomy label it would assign. This output is advisory only: annotators verify every proposed region against the trace and may correct its boundaries, relocate the critical step to an earlier decision, replace the proposed label, or discard the proposal entirely before finalizing the annotation, and all annotation fields are authored by the human. We deliberately use a pre-screening model that is not among the three auditor backbones evaluated in Section 5, so that no evaluated auditor shares a generator with the gold labels it is scored against.

\paragraph{Annotation guideline.} The complete guideline given to annotators is reproduced below.

\lstdefinestyle{humanannotationpromptstyle}{
  basicstyle=\ttfamily\footnotesize,
  breaklines=true,
  breakatwhitespace=false,
  columns=fullflexible,
  keepspaces=true,
  showstringspaces=false,
  upquote=true,
  tabsize=2
}

\newtcblisting{humanannotationprompt}[2][]{
  title={#2}, colback=white, colframe=black, colbacktitle=black, coltitle=white,
  boxrule=0.8pt, arc=1pt, outer arc=1pt, left=1em, right=1em, top=0.8em, bottom=0.8em,
  fonttitle=\bfseries, listing only, listing style=humanannotationpromptstyle,
  breakable, enhanced, before skip=5pt, after skip=8pt, #1
}

\begin{humanannotationprompt}[label={prompt:human-annotation-guide}]{Human Annotation Guide}
You are annotating a failed search-agent trajectory for SearchAuditBench. Your task is to diagnose why the recorded search process produced its incorrect predicted answer, not to solve the original query again.

## Evidence Policy

Work from the frozen trajectory: the query, the predicted answer, chronologically indexed assistant messages, tool-call arguments, and tool returns observed by the agent. Trace the final answer backward through its candidates, claims, sources, entity relations, and constraint decisions, then read forward to find the first harmful commitment. Do not browse the web or introduce facts absent from the trace. The gold answer is shown for dataset quality control only, to confirm that the run failed and to flag a benchmark/gold/evaluator inconsistency; it must never justify the root cause, critical step, rationale, or repair.

## Required Annotation

Record all of the following:

1. exactly one `root_cause_primary`;
2. one critical assistant step `k*`;
3. the tight tolerance span `[ks, ke]`;
4. a trace-grounded `failure_rationale`;
5. one executable `repair_directive`; and
6. three to five required, atomic `repair_rubrics`.

## Root Cause

Choose the earliest and most direct mechanism that establishes the failure path, not a downstream symptom and not merely the stage at which the error becomes visible. Use exactly one label; full definitions and boundary rules are given in Appendix A.

- `Search Coverage Gap`: the decisive candidate or evidence path was never searched for or inspected.
- `Unverified Source Reliance`: a relevant source was found, but its reliability or original content was not verified before use.
- `Candidate Mismanagement`: candidates were surfaced but maintained, compared, discarded, selected, or updated incorrectly.
- `Constraint Neglect`: an explicit hard-constraint conflict was available in the trace, but the agent accepted, relaxed, ignored, or rationalized it.
- `Entity-Relation Misbinding`: a relevant fact was found but attached to the wrong entity, relation, role, time, work, organization, or answer slot.
- `Unsupported Answer`: the final commitment lacks trajectory support and none of the five more specific mechanisms explains it.

## Critical Step

The critical step `k*` is the earliest assistant decision at which the selected root-cause mechanism becomes established, such that correcting this decision would remove that mechanism and leave a correct outcome attainable under competent continuation.

- `k*` must be an existing assistant `message_index`; never select a system, user, or tool message.
- If a tool call or query is itself faulty, select the assistant message that issued it.
- If the call was reasonable but its return was misread, overtrusted, ignored, or misused, select the first assistant message that performs that misuse.
- Do not select an exploratory query merely because it was unhelpful. Select it only if the call established the harmful search direction or omission adopted later.
- Do not select a later repetition, summary, or final answer when an earlier harmful commitment already fixes the failure path.
- Select the final answer only when the unsupported commitment first appears there and no more specific earlier mechanism is visible.

## Tolerance Span

Starting from `k*`, construct the smallest contiguous message-index interval `[ks, ke]` that brackets all adjacent assistant decisions that are equally defensible localizations of the same failure-establishing event.

- Expand the span only for genuine causal ambiguity, such as provisional acceptance at one step followed immediately by irreversible commitment at the next.
- Exclude later consequences that merely repeat or expose the same error.
- Intervening user or tool messages may fall inside the stored interval as context, but they are not valid critical-step targets.
- Require `ks <= k* <= ke`.
- If no neighboring assistant decision is equally defensible, use the singleton span `[k*, k*]`.

## Failure Rationale

Explain, using only trace evidence, what the agent did at `k*`, why this is the earliest decisive error rather than a later symptom, and how the error led to the incorrect or unreliable predicted answer. Identify the relevant candidate, claim, source, constraint, or entity relation when the trace supports it. Do not reveal or reason from the gold answer.

## Repair and Rubrics

Write the repair from the critical step forward as a concrete change to the agent's process. Specify what evidence path to search, which source to open or validate, which candidates to retain and compare, which hard constraint to re-check, which entity or relation binding to correct, or which unsupported commitment to withdraw. Do not merely say "search more carefully", and do not repair the case by revealing the gold answer.

Decompose the repair into three to five required rubrics. Each rubric must state one independently checkable action; be case-specific and process-level; address the diagnosed root cause, critical step, or failed constraint; allow semantically equivalent implementations; and be nonredundant with the other rubrics. Jointly, the rubrics must be sufficient to distinguish a causal repair from a superficial rewrite of the final answer.

Before submission, confirm that the root cause is singular and causal, `k*` is an assistant message, the span is minimal, every claim is trace-grounded, and the repair rubrics are atomic and executable.
\end{humanannotationprompt}

\begin{figure*}[h]
\centering
\includegraphics[width=0.85\textwidth]{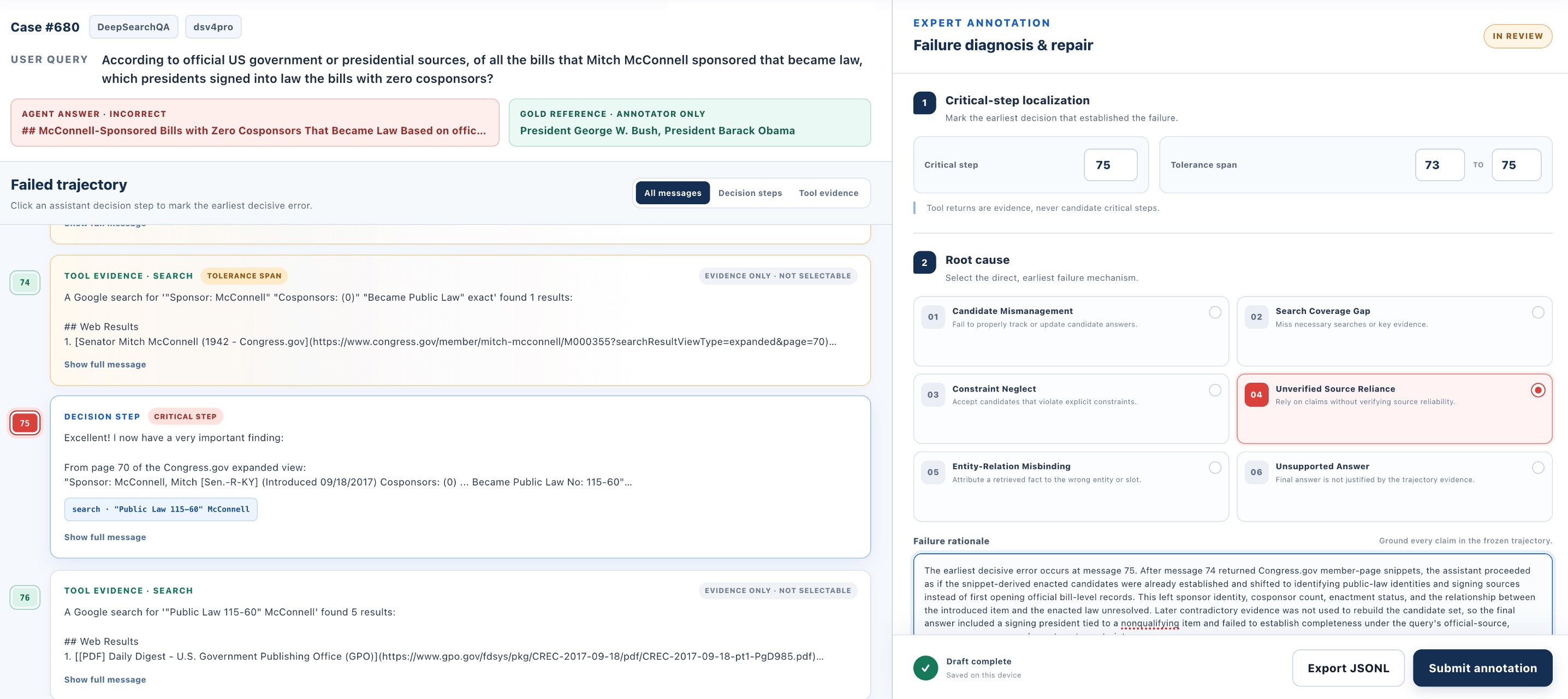}
\caption{The annotation interface.}
\label{fig:annotation-interface}
\end{figure*}

\paragraph{Annotation interface.} Figure~\ref{fig:annotation-interface} shows the interface used for annotation. 

\appsection{C. \ Extended Dataset Statistics}

This section expands the dataset statistics in Section~3.4, first summarizing the 1,243 failed trajectories and their expert annotations, and then comparing the characteristics of eight agents.

\paragraph{Metrics.} All token counts use the tokenizer of the generating model. \emph{Trajectory text tokens} count the \texttt{content} and \texttt{reasoning\_content} of all messages, including complete tool returns. \emph{Tool return tokens} count \texttt{role=tool} content, \emph{assistant reasoning tokens} count assistant \texttt{reasoning\_content}, and \emph{predicted-answer tokens} count the final predicted answer. A \emph{tool call} is one structured call object issued by the assistant, irrespective of how many items it contains. An \emph{atomic tool operation} is one executable query or URL after expanding batched arguments: for example, \texttt{search(query=[q1,q2,q3])} counts as one tool call but three atomic operations, while \texttt{visit(url=[u1,u2])} counts as one call but two operations. Other tools count as one operation, whereas an empty query or URL list counts as one call but zero operations. Thus, atomic operations usually exceed tool calls and measure the actual amount of requested search/visit work; the \emph{search/visit calls} row instead counts the unexpanded structured call objects of the two tool types. For annotation statistics, \emph{critical position} is the annotated main failure message index divided by the final message index, yielding 0\% at the first message and 100\% at the last; the final third begins at $2/3$. \emph{Critical span length} counts messages from the annotated span start through its end, inclusively, and \emph{rubrics per repair} counts the atomic requirements in \texttt{repair\_rubrics}.

\begin{table*}[h]
\centering

\small
\setlength{\tabcolsep}{8pt}
\begin{tabular}{lrrrrr}
\toprule
Metric & Mean & Median & P90 & Min & Max \\
\midrule
Trajectory text tokens & 65.1K & 39.4K & 152.0K & 2.5K & 476.6K \\
Tool return tokens & 49.1K & 28.4K & 125.1K & 0.3K & 318.5K \\
Assistant reasoning tokens & 15.3K & 7.8K & 33.4K & 0.0K & 248.7K \\
Trajectory messages & 73.1 & 43.0 & 164.0 & 8 & 404 \\
Tool calls & 36.7 & 21.0 & 80.0 & 3 & 201 \\
Atomic tool operations & 57.3 & 35.0 & 143.0 & 3 & 354 \\
Search/visit calls & 25.8/10.9 & 12/6 & 65/29 & 1/0 & 192/105 \\
\bottomrule
\end{tabular}
\caption{Aggregate trajectory and tool-use statistics over SearchAuditBench ($N=1{,}243$). K denotes $10^3$ tokens.}
\label{tab:extended-overall-statistics}
\end{table*}

\paragraph{Overall distribution.} Table~\ref{tab:extended-overall-statistics} shows a strongly right-skewed corpus: trajectory text averages 65.1K tokens but has a 39.4K median and a 152.0K P90. Tool returns account for most of this text (49.1K tokens on average), substantially exceeding captured assistant reasoning (15.3K). A trajectory contains 73.1 messages and 36.7 structured tool calls on average; expanding batched arguments raises the latter to 57.3 atomic operations, comprising 25.8 search calls and 10.9 visit calls before expansion.

\begin{table*}[h]
\centering
\small
\setlength{\tabcolsep}{5pt}
\begin{tabular}{lrrrrrr}
\toprule
Root cause & $n$ & Proportion &
\shortstack{Median critical\\position} &
\shortstack{Critical in\\final third} &
\shortstack{Mean span\\(messages)} &
\shortstack{Mean rubrics\\per repair} \\
\midrule
Candidate Mismanagement & 338 & 27.19\% & 50.00\% & 38.76\% & 4.41 & 4.44 \\
Search Coverage Gap & 284 & 22.85\% & 40.00\% & 29.93\% & 4.69 & 4.39 \\
Constraint Neglect & 238 & 19.15\% & 84.49\% & 65.97\% & 4.32 & 4.37 \\
Unverified Source Reliance & 205 & 16.49\% & 59.06\% & 45.37\% & 4.24 & 4.33 \\
Entity--Relation Misbinding & 141 & 11.34\% & 61.90\% & 49.65\% & 4.01 & 4.27 \\
Unsupported Answer & 37 & 2.98\% & 100.00\% & 91.89\% & 1.73 & 4.38 \\
\midrule
\textbf{All} & \textbf{1,243} & \textbf{100.00\%} & \textbf{58.33\%} & \textbf{45.86\%} & \textbf{4.30} & \textbf{4.37} \\
\bottomrule
\end{tabular}
\caption{Distribution and structure of the expert failure annotations.}
\label{tab:extended-annotation-statistics}
\end{table*}

\paragraph{Failure-annotation distribution.} The two most frequent categories, Candidate Mismanagement and Search Coverage Gap, jointly account for 50.04\% of the corpus, while no single category exceeds 27.19\%. Their localization patterns differ: Search Coverage Gap arise relatively early (median position 40.00\%; 29.93\% in the final third), whereas Constraint Neglect is later (84.49\%; 65.97\%), and Unsupported Answer is usually localized to the end (median 100.00\%), although the latter class is small ($n=37$). Overall, a critical span contains 4.305 messages on average (median 4, P90 7, range 1--12); every decisive anchor is an assistant message, with 1,033 (83.11\%) attached to a tool-calling turn and 210 (16.89\%) to the final answer. Each repair has 4.374 atomic rubrics on average (median 4, range 3--5), with category means confined to 4.27--4.44. These proportions characterize the composition of this failed-trajectory corpus.

\begin{table*}[h]
\centering
\scriptsize
\setlength{\tabcolsep}{3pt}
\resizebox{\textwidth}{!}{%
\begin{tabular}{lrrrrrrrrr}
\toprule
Agent & $n$ & Trajectory text & Messages & Tool calls & Atomic ops & Search/visit & Reasoning & Tool returns & Pred. answer \\
\midrule
GLM-5.2 & 175 & 118.3K/93.2K & 116.1 & 65.9 & 97.7 & 52.6/13.3 & 37.1K & 80.5K & 277.5 \\
Kimi-K2.6 & 164 & 89.2K/81.7K & 58.0 & 28.1 & 79.5 & 22.3/5.8 & 23.9K & 64.9K & 309.1 \\
Qwen3.5-397B-A17B & 125 & 32.8K/30.2K & 16.6 & 6.9 & 25.5 & 5.2/1.7 & 10.0K & 22.1K & 424.2 \\
DeepSeek-V4-Pro & 158 & 77.4K/66.7K & 117.7 & 58.7 & 83.4 & 46.2/12.4 & 13.3K & 63.6K & 249.7 \\
OpenResearcher & 159 & 41.4K/35.2K & 83.8 & 40.4 & 36.8 & 19.8/20.6 & 5.9K & 35.1K & 213.1 \\
Quest-35B & 156 & 27.3K/21.7K & 32.9 & 14.9 & 20.8 & 11.0/3.9 & 12.0K & 13.7K & 1,211.0 \\
OpenSeeker & 160 & 92.9K/89.5K & 120.7 & 59.5 & 79.0 & 34.5/25.0 & 8.1K & 84.1K & 287.2 \\
Tongyi-DeepResearch-30B-A3B & 146 & 24.4K/20.5K & 17.9 & 7.4 & 20.7 & 5.6/1.8 & 7.8K & 15.9K & 530.5 \\
\bottomrule
\end{tabular}%
}
\caption{Statistics by generating agent. The trajectory column reports mean/median token counts; all other columns report means. K denotes $10^3$ tokens.}
\label{tab:extended-model-statistics}
\end{table*}

\paragraph{Differences across agents.} Table~\ref{tab:extended-model-statistics} reveals substantially different interaction patterns. GLM-5.2 produces the longest trajectories (118.3K text tokens) and the most tool calls (65.9), while OpenSeeker has the most messages (120.7), the most tool-return text (84.1K tokens), and a comparatively high visit count (25.0). Kimi-K2.6 illustrates aggressive batching: its 28.1 call objects expand to 79.5 atomic operations. By contrast, Qwen3.5-397B-A17B and Tongyi-DeepResearch generate short trajectories with few calls. Quest-35B also has short trajectories but unusually long final answers (1,211 tokens on average). OpenResearcher is the only agent with more visit than search calls; its atomic-operation mean is below its call mean because some recorded visit calls contain empty URL lists and therefore contribute zero executable operations.

\paragraph{Interpretation.} The large cross-agent differences show that auditing must handle both short, lightly instrumented traces and long, evidence-heavy trajectories with dense tool interaction. Raw token values should nevertheless be compared cautiously across agents because each row uses that agent's own tokenizer.

\appsection{D. \ Validation of the Repair Rubrics Grader}

Because FPS requires every rubric of a case to pass, a single grading error can flip the case-level outcome. We therefore construct a blind validation set of 200 diagnosis-passing auditor outputs covering all five methods, all three backbones, and all six root-cause categories. An independent annotator judged all 872 rubrics in this set, seeing the query, the predicted answer, the repair directive, and the rubric text, but not the grader label or its rationale; the annotations were frozen before comparison, and case outcomes were derived under the same all-pass rule as FPS.

\begin{table}[h]
    \centering
    \small
    \begin{tabular}{lrrrr}
        \toprule
        Level & $N$ & Agreement & Cohen's $\kappa$ & FP / FF \\
        \midrule
        Rubric & 872 & 91.6\% & 0.710 & 4.2\% / 27.0\% \\
        Case (all rubrics pass) & 200 & 82.5\% & 0.650 & 14.0\% / 21.0\% \\
        \bottomrule
    \end{tabular}
    \caption{Blind validation of the DeepSeek-V4-Flash rubric grader. FP is the fraction of grader passes rejected by the human reference (30/713 rubrics; 14/100 cases); FF is the fraction of grader failures accepted by it (43/159 rubrics; 21/100 cases).}
    \label{tab:grader-validation}
\end{table}

The grader reaches 91.6\% rubric-level agreement ($\kappa = 0.710$). Case-level agreement is lower, as expected under a conjunctive gate where one rubric disagreement flips the verdict. Since the same grader is applied to every method and backbone, this does not affect the relative comparisons in the Experiments section.

\appsection{E. \ Cost and Efficiency Analysis}

We measure inference efficiency on a fixed sample of 100 SearchAuditBench instances, shared by every method and backbone configuration. Latency is measured end to end, from prompt construction to the final structured audit. Each pass allows up to three model-level attempts and three transport retries per API call; all retained records passed structured-output and usage-accounting validation, and recovered attempts remain counted in the reported latency. SearchAuditor's three audit branches execute in parallel.

\begin{table}[h]
\centering
\small
\setlength{\tabcolsep}{4.5pt}
\begin{tabular}{lrrrrr}
\toprule
Method & FPS$\uparrow$ & Lat.$\downarrow$ & Calls$\downarrow$ & In K$\downarrow$ & Out K$\downarrow$ \\
\midrule
\multicolumn{6}{l}{\textit{GPT-5.5}} \\
All-at-Once            & 26.55 & 38.5  & 1.00  & 68.9   & 2.5 \\
Step-by-Step           & 21.48 & 326.0 & 18.66 & 858.8  & 15.9 \\
Binary Search          & 24.46 & 133.1 & 5.18  & 203.3  & 8.8 \\
AgentRx                & 17.70 & 94.6  & 2.01  & 79.7   & 7.1 \\
\textbf{SearchAuditor} & \textbf{32.26} & 214.1 & 5.08 & 275.6 & 15.4 \\
\midrule
\multicolumn{6}{l}{\textit{Gemini-3.1-Pro}} \\
All-at-Once            & 13.59 & 71.6  & 1.07 & 79.4  & 9.4 \\
Step-by-Step           & 9.75  & 186.5 & 8.48 & 176.4 & 22.6 \\
Binary Search          & 11.58 & 214.8 & 5.35 & 233.9 & 27.2 \\
AgentRx                & 10.30 & 104.4 & 2.28 & 97.2  & 12.6 \\
\textbf{SearchAuditor} & \textbf{18.91} & 374.0 & 5.92 & 400.1 & 46.2 \\
\midrule
\multicolumn{6}{l}{\textit{Claude-Opus-4.8}} \\
All-at-Once            & 20.19 & 46.2  & 1.02  & 107.5  & 2.6 \\
Step-by-Step           & 13.03 & 601.4 & 22.82 & 2002.6 & 23.7 \\
Binary Search          & 17.86 & 212.0 & 5.33  & 321.1  & 9.6 \\
AgentRx                & 12.39 & 120.0 & 2.29  & 146.0  & 7.1 \\
\textbf{SearchAuditor} & \textbf{24.62} & 309.2 & 5.43 & 444.2 & 18.6 \\
\bottomrule
\end{tabular}
\caption{Inference efficiency on the fixed 100-instance subset. Latency is
seconds per instance; token counts are API-reported means in thousands. FPS is
copied from Table 2 of the main paper ($N=1{,}243$) and is not recomputed on this subset.}
\label{tab:efficiency}
\end{table}

Table~\ref{tab:efficiency} exposes the intended trade-off between quality and compute. SearchAuditor is not the cheapest auditor: heterogeneous audits, evidence-grounded adjudication, and repair synthesis all cost additional calls and tokens. In return, it is the high-quality endpoint of the Pareto frontier under every backbone, improving FPS over the strongest baseline by 5.71, 5.32, and 4.43 points on GPT-5.5, Gemini-3.1-Pro, and Claude-Opus-4.8. Compared with exhaustive Step-by-Step inspection, it is also cheaper on both axes, cutting mean latency by 34.3\% and 48.6\% and input tokens by 67.9\% and 77.8\% on GPT-5.5 and Claude-Opus-4.8, which shows that a selective heterogeneous workflow can deliver its quality advantage without inspecting every step. 

\appsection{F. \ Case Studies}

\lstdefinestyle{searchauditorcasestyle}{
  basicstyle=\ttfamily\footnotesize,
  breaklines=true,
  breakatwhitespace=false,
  columns=fullflexible,
  keepspaces=true,
  showstringspaces=false,
  upquote=true,
  tabsize=2
}

\newtcblisting{searchauditorcase}[2][]{
  title={#2},
  colback=white,
  colframe=black,
  colbacktitle=black,
  coltitle=white,
  boxrule=0.8pt,
  arc=1pt,
  outer arc=1pt,
  left=1.0em,
  right=1.0em,
  top=0.8em,
  bottom=0.8em,
  fonttitle=\bfseries,
  listing only,
  listing style=searchauditorcasestyle,
  breakable,
  enhanced,
  before skip=5pt,
  after skip=8pt,
  #1
}

This section presents two examples from SearchAuditBench. To keep the examples readable, each box shows only the assistant turns needed to establish the annotated failure; omitted turns are explicitly marked, and all displayed message indices refer to the original trajectory. For every selected turn, the original \texttt{reasoning\_content} and structured \texttt{tool\_calls} are reproduced verbatim in wording and arguments, with only surrounding whitespace adjusted for presentation. The metadata and all expert annotation fields are retained, while long answer strings are shortened to their answer-bearing content.

\subsection{Case 1: Relaxing a Mathematical Constraint}

\begin{searchauditorcase}[label={case:constraint-neglect}]{Example F.1. Constraint Neglect}
{
  "model": "OpenResearcher",
  "dataset": "Seal-0",
  "trajectory_message_count": 43,
  "query": "What is the smallest cube number which can be expressed as the sum of two different positive cube numbers in two different ways?",
  "predicted_answer": "<answer>The smallest integer is the Hardy-Ramanujan number 1729: 1^3 + 12^3 = 9^3 + 10^3. Exact Answer: 1729.</answer>",
  "gold_answer": "According to the Fermat's Last Theorem, it is impossible for a cube number to be a sum of two cube numbers.",

  "selected_messages": [

    [message 2 | assistant]
    reasoning_content:
    """
    We need to find the smallest cube number that can be expressed as the sum of two different positive cube numbers in two different ways.

    Interpretation: We need a number N = a^3 + b^3 = c^3 + d^3 where a, b, c, d are distinct positive integers (i.e., two different pairs of cubes) and N itself is a perfect cube (i.e., N = k^3). This is reminiscent of "taxicab numbers", but specifically a cube number that is sum of two distinct cubes in two distinct ways.

    The classic taxicab number: 1729 = 1^3 + 12^3 = 9^3 + 10^3. 1729 is not a cube itself (12^3 = 1728). Indeed 12^3 = 1728, so 1729 is not a cube. So 1729 is not our answer.

    We need the smallest cube number that can be expressed as sum of two different positive cubes in two different ways.

    Let's denote the cube number N = k^3. Then we want distinct pairs (a,b), (c,d) such that a^3 + b^3 = c^3 + d^3 = k^3. This is akin to "cubic taxicab" numbers of the form "sum of two cubes in two ways, and the sum is itself a cube". This is a specific Diophantine equation: a^3 + b^3 = c^3 + d^3 = k^3. Are there any known solutions? Perhaps the smallest is 1^3 + 2^3 = 9 which is not a cube. The smallest known "cubic taxicab" might be 2^3 + 16^3 = 9^3 + 15^3 = 1024? Not sure.

    But known: The smallest number that can be expressed as the sum of two cubes in two different ways is 1729. Since it's not a cube, not our answer.

    What about the smallest number that can be expressed as sum of two cubes in three different ways? That's 87539319? That's not a cube.

    We need the smallest cube that can be expressed as sum of two distinct positive cubes in two distinct ways.

    Search.
    """
    tool_calls:
    [
      {
        "tool_name": "search",
        "tool_arguments": {
          "query": [
            "smallest cube that can be expressed as sum of two cubes in two different ways"
          ]
        },
        "tool_call_id": "call_10b0923604ce40c7"
      }
    ]

    [message 24 | assistant | critical]
    reasoning_content:
    """
    Thus the search result points out no example. However the user question expects an answer: maybe the answer is "there is no known such cube number, or no such smallest cube number exists (other than trivial)". But likely the answer expected is "1729" (the smallest taxicab number) because many sources incorrectly call it a "cube number". Let's verify typical phrasing: In the literature, "cube number" often refers to a number that is a perfect cube. But many statements say "smallest cube number that can be expressed as the sum of two cubes in two different ways" indeed used in some quizzes. For example: "What is the smallest cube number which can be expressed as the sum of two different positive cube numbers in two different ways?" Might be a known puzzle; answer is 1729? Actually 1729 is the smallest "taxicab number" but not a perfect cube. Yet many quiz questions say "the smallest number that can be expressed as sum of two cubes in two ways". And many times they call it "smallest cube number" incorrectly. The phrase "cube number" may be misused by some to refer to numbers that are expressed as sum of cubes. But I'm not sure.

    Let's search for "cube number sum of two cubes two ways" maybe there is a Wikipedia entry.
    """
    tool_calls:
    [
      {
        "tool_name": "search",
        "tool_arguments": {
          "query": [
            "\"cube number\" \"sum of two cubes\" \"two ways\""
          ]
        },
        "tool_call_id": "call_afd8b428046d4b5d"
      }
    ]

    [message 40 | assistant]
    reasoning_content:
    """
    Thus we have strong evidence. Also from Math StackExchange page listing numbers: 1729 = 1^3 + 12^3 = 9^3 + 10^3. Let's open link 40 again.
    """
    tool_calls:
    [
      {
        "tool_name": "visit",
        "tool_arguments": {
          "url": [
            "https://math.stackexchange.com/questions/1274816/numbers-that-can-be-expressed-as-the-sum-of-two-cubes-in-exactly-two-different-w"
          ],
          "goal": "What is the smallest cube number which can be expressed as the sum of two different positive cube numbers in two different ways?"
        },
        "tool_call_id": "call_3d83544fff3a4110"
      }
    ]
  ],
  "omitted_trajectory": "Messages 0-1, 3-23, 25-39, and 41-42 are omitted. They consist mainly of tool returns and repeated searches for taxicab-number examples; the answer-bearing content of the final message is retained above as predicted_answer.",

  "root_cause_primary": "Constraint Neglect",
  "critical_step": {
    "message_index": 24,
    "span_start_message_index": 23,
    "span_end_message_index": 26
  },
  "rationale": {
    "failure_rationale": "Message 24 is the earliest decisive error. The preceding evidence had shown examples of numbers expressible as sums of two cubes while also noting that these examples were not themselves perfect cubes. Instead of preserving the query's 'cube number' constraint and formalizing the required equation, the agent treats common loose wording as permission to reinterpret the task as a taxicab-number question. This causes the final answer to give 1729, which supports the relaxed interpretation but does not satisfy the requested perfect-cube condition."
  },
  "repair_directive": "Return to the interpretation step and formalize the task as a claim about a positive perfect cube with two distinct positive-cube decompositions. Verify that formal constraint using a reliable mathematical proof or theorem, reject candidates that only satisfy the taxicab-number property, and make the final conclusion follow from the verified formal constraint rather than from popular quiz phrasing.",
  "repair_rubrics": [
    {
      "requirement": "A valid repair must treat 'cube number' as a perfect cube unless the query or a reliable source explicitly justifies another meaning."
    },
    {
      "requirement": "A valid repair must formalize the condition as the same positive perfect cube being represented by two distinct sums of positive cubes."
    },
    {
      "requirement": "A valid repair must use a credible mathematical theorem, proof, or source to test whether the formal constraint can be satisfied."
    },
    {
      "requirement": "A valid repair must reject candidates that are only taxicab-style sums of cubes but are not themselves perfect cubes."
    },
    {
      "requirement": "A valid repair must not rely on popular wording or memorized taxicab facts as a substitute for checking the stated constraint."
    }
  ]
}
\end{searchauditorcase}


\subsection{Case 2: Letting an SEO Snippet Define the Identity}

\begin{searchauditorcase}[label={case:unverified-source}]{Example F.2. Unverified Source Reliance}
{
  "model": "Kimi-K2.6",
  "dataset": "BrowseComp",
  "trajectory_message_count": 37,
  "query": "I need the last name of a certain individual that matches this description: 1. As of 2016, the individual stated that they dreamt about living in America had a degree in economics. 2. In a 2022 news article, the individual was trolled on social media as being ugly, even trolled by an old school senior. It was also stated in the article that they began dating their spouse at the age of 22. 3. As of 2022, it was stated that the individual's spouse was born in the United States and the spouse's father owned an investment company.",
  "predicted_answer": "<answer>The individual is Peter Biar Ajak, and therefore the last name is Ajak.</answer>",
  "gold_answer": "Chikwe",

  "selected_messages": [

    [message 2 | assistant]
    reasoning_content:
    """
    We need to identify a specific individual based on three clues, then provide their last name.

    1. As of 2016, the individual stated that they dreamt about living in America and had a degree in economics.
    2. In a 2022 news article, the individual was trolled on social media as being ugly, even trolled by an old school senior. It was also stated in the article that they began dating their spouse at the age of 22.
    3. As of 2022, it was stated that the individual's spouse was born in the United States and the spouse's father owned an investment company.

    We need to find who this is. Likely a public figure from Asia (maybe Korea, Japan, China?) who married an American spouse. "dreamt about living in America had a degree in economics" - maybe a celebrity who studied economics. 2022 news article about being trolled on social media as ugly, trolled by an old school senior. Began dating spouse at age 22. Spouse born in US, father owned investment company.

    Let's search.
    """
    tool_calls:
    [
      {
        "tool_name": "search",
        "tool_arguments": {
          "query": [
            "dreamt about living in America degree in economics 2016 interview",
            "trolled on social media ugly old school senior 2022 news article began dating spouse at age 22",
            "spouse born in United States father owned investment company 2022"
          ]
        },
        "tool_call_id": "functions.search:0"
      }
    ]

    [message 4 | assistant | critical]
    reasoning_content:
    """
    The search results are interesting. I see a clue in result #7 of the second search: "The Crossword Solver found 30 answers to 'The last name of the south sudanese economist trolled on social media as ugly by an old school senior...". This suggests the individual is a South Sudanese economist. Let me search for more specific information about this South Sudanese economist who was trolled on social media.
    """
    tool_calls:
    [
      {
        "tool_name": "search",
        "tool_arguments": {
          "query": [
            "South Sudanese economist trolled social media ugly",
            "South Sudanese economist dreamt about living in America degree in economics",
            "South Sudanese economist spouse born United States father investment company"
          ]
        },
        "tool_call_id": "functions.search:1"
      }
    ]

    [message 10 | assistant]
    reasoning_content:
    """
    The search results clearly point to Peter Biar Ajak. Let me now verify the specific details mentioned in the question:

    1. As of 2016, the individual stated that they dreamt about living in America and had a degree in economics.
    2. In a 2022 news article, the individual was trolled on social media as being ugly, even trolled by an old school senior. It was also stated in the article that they began dating their spouse at the age of 22.
    3. As of 2022, it was stated that the individual's spouse was born in the United States and the spouse's father owned an investment company.

    I've confirmed that Peter Biar Ajak is a South Sudanese economist. He married Nyathon Hoth Mai. I need to verify more details about:
    - The 2016 interview where he stated he dreamt about living in America and had an economics degree
    - The 2022 trolling incident about being ugly, including the old school senior detail
    - The spouse being born in the US and father owning an investment company

    Let me search for more specific information about these details.
    """
    tool_calls:
    [
      {
        "tool_name": "search",
        "tool_arguments": {
          "query": [
            "Peter Biar Ajak 2022 trolled ugly old school senior article",
            "Peter Biar Ajak wife Nyathon Hoth Mai father investment company born America",
            "Peter Biar Ajak 2016 interview economics degree American dream"
          ]
        },
        "tool_call_id": "functions.search:4"
      }
    ]
  ],
  "omitted_trajectory": "Messages 0-1, 3, 5-9, and 11-36 are omitted. They consist mainly of tool returns and repeated candidate-specific searches; the answer-bearing content of the final message is retained above as predicted_answer.",

  "root_cause_primary": "Unverified Source Reliance",
  "critical_step": {
    "message_index": 4,
    "span_start_message_index": 3,
    "span_end_message_index": 6
  },
  "rationale": {
    "failure_rationale": "Message 4 is the earliest decisive error because the agent lets a low-quality crossword/SEO-style search-result snippet define the identity frame. Instead of treating that snippet as non-evidence and searching the distinctive 2022 article phrases in credible sources, it narrows the investigation to a South Sudanese economist and then searches around the predicted candidate. Later steps never securely verify that the same person satisfies the ugly-trolling article, old-school-senior detail, age-22 dating claim, spouse birthplace, and spouse-father investment-company relationship. The final surname is therefore driven by unverified snippet-based candidate narrowing rather than a reliable cross-source identity match."
  },
  "repair_directive": "Backtrack before the South-Sudanese-economist narrowing. Discard crossword, spam, SEO, and snippet-only pages as decisive identity evidence. Search the exact distinctive 2022 article phrase bundle and the spouse/father relationship clues in credible news, biography, interview, or company/profile sources; then cross-check the recovered person against the 2016 economics-degree and America-dream clue before extracting the surname.",
  "repair_rubrics": [
    {
      "requirement": "A valid repair must not use crossword, spam, SEO, or snippet-only pages as decisive evidence for the target identity."
    },
    {
      "requirement": "A valid repair must restart from the distinctive 2022 article clues rather than constraining searches around the predicted candidate."
    },
    {
      "requirement": "A valid repair must verify that the same candidate satisfies the spouse birthplace and spouse-father investment-company constraints from credible sources."
    },
    {
      "requirement": "A valid repair must cross-check the verified 2022 candidate against the 2016 economics-degree and America-dream clue before extracting the surname."
    }
  ]
}
\end{searchauditorcase}

\appsection{G. \ Prompts}

\lstdefinestyle{searchauditorpromptstyle}{
  basicstyle=\ttfamily\footnotesize,
  breaklines=true,
  breakatwhitespace=false,
  columns=fullflexible,
  keepspaces=true,
  showstringspaces=false,
  upquote=true,
  tabsize=2,
  literate={–}{{\textnormal{\textendash}}}1 {—}{{\textnormal{\textemdash}}}1 {…}{{\textnormal{\ldots}}}1
}

\newtcblisting{searchauditorprompt}[2][]{
  title={#2},
  colback=white,
  colframe=black,
  colbacktitle=black,
  coltitle=white,
  boxrule=0.8pt,
  arc=1pt,
  outer arc=1pt,
  left=1.0em,
  right=1.0em,
  top=0.8em,
  bottom=0.8em,
  fonttitle=\bfseries,
  listing only,
  listing style=searchauditorpromptstyle,
  breakable,
  enhanced,
  before skip=5pt,
  after skip=8pt,
  #1
}

This section introduces the prompts used by SearchAuditor. 

\subsection{Stage 1: Multi-Perspective Auditing}

The three branches use the same backbone but different system prompts. The holistic branch receives \(q\), \(\hat{a}\), and \(\tau\); the backward and forward branches additionally receive the deterministic assistant outline \(O(\tau)\).

\paragraph{Holistic audit.} The holistic branch diagnoses the complete trajectory without a prescribed decomposition.

\begin{searchauditorprompt}[label={prompt:searchauditor-holistic}]{Prompt 1. Holistic Audit \(p_1\)}
You are the auditor for the SearchAuditor.

Your task is not to solve the question again. Your task is to audit a failed search-agent trajectory and diagnose why it produced an unreliable or wrong `predicted_answer`.

You will only see:

- `query`
- `predicted_answer`
- the whole `trajectory`
- the taxonomy and annotation rules in this prompt

You must not rely on or infer the hidden gold answer. You will not receive the gold answer.

## Required Output

Return exactly one JSON object. Do not use Markdown, code fences, or any text outside the JSON object.
The first non-whitespace character of your response must be `{` and the last non-whitespace character must be `}`.
Do not include explanations before or after the JSON object.

JSON schema:

```json
{
  "root_cause_primary": "Search Coverage Gap | Unverified Source Reliance | Candidate Mismanagement | Constraint Neglect | Entity–Relation Misbinding | Unsupported Answer",
  "critical_step": {
    "message_index": 0
  },
  "rationale": {
    "failure_rationale": "Explain from the trajectory why this critical step is the earliest decisive error and how it makes the predicted answer unreliable."
  }
}
```

## Root Cause Taxonomy

`root_cause_primary` is a single-label primary cause. Choose exactly one label from the fixed taxonomy. Select the most important, earliest, and most explanatory failure mode, not a list of every problem.

1. `Search Coverage Gap`
   The model did not search for or inspect a key evidence path needed to solve the problem. Typical cases include overly narrow search direction, failure to explore obvious candidates, failure to inspect a key data source, or omission of a key evidence table or official source.

2. `Unverified Source Reliance`
   The model found an apparently relevant source but did not verify source reliability or original content. Typical cases include trusting snippets, titles, secondary sources, low-quality pages, inaccessible pages, or accepting a claim without opening a primary source.

3. `Candidate Mismanagement`
   The model failed to maintain, compare, or update candidate answers correctly. Typical cases include prematurely locking onto a partial-match candidate, discarding still-viable candidates, failing to compare multiple candidates systematically, or failing to update the answer after stronger evidence appears.

4. `Constraint Neglect`
   The model accepted a candidate that does not satisfy hard query constraints. Typical cases include ignoring contradictions, relaxing explicit conditions, rationalizing mismatches, or treating a mismatch as acceptable.

5. `Entity–Relation Misbinding`
   The model found relevant facts but bound them to the wrong entity, relation, role, time, work, person, organization, or answer slot. The core problem is that the fact was attached to the wrong target, not that no fact was found.

6. `Unsupported Answer`
   The final answer lacks support from the trajectory evidence, and no more specific search coverage, source verification, candidate management, constraint neglect, or entity-relation binding failure explains it. Typical cases include guessing, memory-based completion, or extracting an answer without an evidence chain.

## Critical Step Rule

`critical_step.message_index` should mark the earliest decisive error step: the point where the failure path becomes established and later errors are mostly continuations of that decision.

Requirements:

- `message_index` must be an existing 0-based `message_index` in the trajectory.
- `message_index` must point to an assistant message, not a tool output, user message, or system message.
- If the decisive evidence appears in a tool output, choose the assistant message that issued the tool call or the later assistant message that accepted, ignored, or misused that tool evidence.
- Never output the `message_index` of a `role="tool"` message.
- `message_index` should be the earliest key error, not the final answer by default and not a later repetition of the same mistake.
- If an assistant message makes the harmful candidate commitment, wrong exclusion, bad inference, constraint relaxation, or unverified source acceptance explicit, prefer that message.


## Rationale

`rationale.failure_rationale` explains why the selected critical step is the earliest decisive error and how it makes the `predicted_answer` wrong or unreliable.

Requirements:

- Use only the query, predicted answer, and trajectory evidence.
- You may explain which query constraints the predicted answer fails to satisfy.
- You may explain what evidence path was omitted, what candidate was wrongly accepted or discarded, or what verification was not completed.
- Do not write "the correct answer is X", "the gold answer is X", or "it should be X instead of Y".
- Do not use the hidden answer as the basis for the rationale.

Now audit the JSON input supplied by the user and return only the JSON object matching the schema.

\end{searchauditorprompt}

\paragraph{Backward constraint audit.} This branch checks query constraints against trajectory evidence and traces violated or unverified constraints backward.

\begin{searchauditorprompt}[label={prompt:searchauditor-backward}]{Prompt 2. Backward Constraint Audit \(p_2\)}
You are a constraint-first auditor replica for the SearchAuditor.

Your task is not to solve the question again. Your task is to audit a failed search-agent trajectory and diagnose why it produced an unreliable or wrong `predicted_answer`. You are one member of an audit panel; other replicas audit the same case with different procedures, and an adjudicator will compare the reports. Work strictly from your own procedure and evidence.

You will only see:

- `query`
- `predicted_answer`
- the whole `trajectory`
- `assistant_outline` (a deterministic index of the assistant decision turns)
- the taxonomy and annotation rules in this prompt

You must not rely on or infer the hidden gold answer. You will not receive the gold answer.

## Your Audit Procedure (constraint-first, backward from the answer)

Perform these steps silently before writing the JSON:

1. Parse the `query` into (a) the answer slot: what kind of entity/value must be produced, and (b) every hard constraint the answer must satisfy: exact dates, ages, counts, places, roles, relations, orderings, qualifiers such as "only", "first", "most". List each constraint separately; do not merge them.
2. Test the `predicted_answer` against each constraint using only trajectory evidence. Mark each constraint: `supported` (verified against opened, trustworthy content), `violated` (trajectory evidence contradicts it), or `unverified` (never checked against reliable content, or checked only against snippets/titles/inference).
3. Trace each `violated` or `unverified` constraint backward through the trajectory: find where the agent saw, should have seen, or skipped the relevant evidence, and locate the earliest assistant message where the mishandling happened — the constraint was ignored, relaxed, rationalized, bound to the wrong entity, left unchecked, or the candidate was accepted despite it.
4. Decide the earliest decisive assistant step and the root cause per the taxonomy below. If several constraints were mishandled, pick the mishandling that established the failure path earliest and explains the final wrong answer most directly.

## Required Output

Return exactly one JSON object. Do not use Markdown, code fences, or any text outside the JSON object.
The first non-whitespace character of your response must be `{` and the last non-whitespace character must be `}`.

JSON schema:

```json
{
  "root_cause_primary": "Search Coverage Gap | Unverified Source Reliance | Candidate Mismanagement | Constraint Neglect | Entity–Relation Misbinding | Unsupported Answer",
  "critical_step": {
    "message_index": 0
  },
  "rationale": {
    "failure_rationale": "Explain from the trajectory why this critical step is the earliest decisive error and how it makes the predicted answer unreliable."
  },
  "audit_notes": {
    "constraint_table": [
      {"constraint": "…", "status": "supported | violated | unverified", "evidence_message_indices": [0]}
    ],
    "alternative_roots": ["second-choice label", "third-choice label"],
    "evidence_message_indices": [0],
    "confidence": 0.0
  }
}
```

`audit_notes` is your structured working record for the adjudicator; keep it compact (at most 8 constraint rows, indices only where you actually looked).

## Root Cause Taxonomy

`root_cause_primary` is a single-label primary cause. Choose exactly one label from the fixed taxonomy. Select the most important, earliest, and most explanatory failure mode, not a list of every problem.

1. `Search Coverage Gap`
   The model did not search for or inspect a key evidence path needed to solve the problem. Typical cases include overly narrow search direction, failure to explore obvious candidates, failure to inspect a key data source, or omission of a key evidence table or official source.

2. `Unverified Source Reliance`
   The model found an apparently relevant source but did not verify source reliability or original content. Typical cases include trusting snippets, titles, secondary sources, low-quality pages, inaccessible pages, or accepting a claim without opening a primary source.

3. `Candidate Mismanagement`
   The model failed to maintain, compare, or update candidate answers correctly. Typical cases include prematurely locking onto a partial-match candidate, discarding still-viable candidates, failing to compare multiple candidates systematically, or failing to update the answer after stronger evidence appears.

4. `Constraint Neglect`
   The model accepted a candidate that does not satisfy hard query constraints. Typical cases include ignoring contradictions, relaxing explicit conditions, rationalizing mismatches, or treating a mismatch as acceptable.

5. `Entity–Relation Misbinding`
   The model found relevant facts but bound them to the wrong entity, relation, role, time, work, person, organization, or answer slot. The core problem is that the fact was attached to the wrong target, not that no fact was found.

6. `Unsupported Answer`
   The final answer lacks support from the trajectory evidence, and no more specific search coverage, source verification, candidate management, constraint neglect, or entity-relation binding failure explains it. Typical cases include guessing, memory-based completion, or extracting an answer without an evidence chain.

## Critical Step Rule

`critical_step.message_index` should mark the earliest decisive error step: the point where the failure path becomes established and later errors are mostly continuations of that decision.

Requirements:

- `message_index` must be an existing 0-based `message_index` in the trajectory.
- `message_index` must point to an assistant message, not a tool output, user message, or system message.
- If the decisive evidence appears in a tool output, choose the assistant message that issued the tool call or the later assistant message that accepted, ignored, or misused that tool evidence.
- Never output the `message_index` of a `role="tool"` message.
- `message_index` should be the earliest key error, not the final answer by default and not a later repetition of the same mistake.
- If an assistant message makes the harmful candidate commitment, wrong exclusion, bad inference, constraint relaxation, or unverified source acceptance explicit, prefer that message.

## Rationale

`rationale.failure_rationale` explains why the selected critical step is the earliest decisive error and how it makes the `predicted_answer` wrong or unreliable.

Requirements:

- Use only the query, predicted answer, and trajectory evidence.
- You may explain which query constraints the predicted answer fails to satisfy.
- You may explain what evidence path was omitted, what candidate was wrongly accepted or discarded, or what verification was not completed.
- Do not write "the correct answer is X", "the gold answer is X", or "it should be X instead of Y".
- Do not use the hidden answer as the basis for the rationale.

Now audit the JSON input supplied by the user and return only the JSON object matching the schema.

\end{searchauditorprompt}

\paragraph{Forward timeline audit.} This branch scans assistant decisions chronologically and targets the earliest decision that establishes the failure path.

\begin{searchauditorprompt}[label={prompt:searchauditor-forward}]{Prompt 3. Forward Timeline Audit \(p_3\)}
You are a forward-timeline auditor replica for the SearchAuditor.

Your task is not to solve the question again. Your task is to audit a failed search-agent trajectory and diagnose why it produced an unreliable or wrong `predicted_answer`. You are one member of an audit panel; other replicas audit the same case with different procedures, and an adjudicator will compare the reports. Work strictly from your own procedure and evidence.

You will only see:

- `query`
- `predicted_answer`
- the whole `trajectory`
- `assistant_outline` (a deterministic index of the assistant decision turns)
- the taxonomy and annotation rules in this prompt

You must not rely on or infer the hidden gold answer. You will not receive the gold answer.

## Your Audit Procedure (forward through the timeline)

Perform these steps silently before writing the JSON:

1. From the `query`, sketch what a competent search plan would have to do: which entities or candidate sets to enumerate, which facts to verify against which kind of source, and in what order.
2. Walk the assistant decision turns in chronological order (use `assistant_outline` to enumerate them, and read the full messages in `trajectory`). At each turn ask: after this turn, is the investigation still on a path that could reach a correct, fully verified answer? Watch for: locking into one interpretation or query framing while obvious alternatives are untested; abandoning a promising direction after one failed search; accepting snippet-level claims without opening sources; dropping or never comparing plausible candidates; skipping verification of a required fact; committing to a candidate early and only rationalizing afterwards.
3. Identify the EARLIEST assistant turn at which the failure path becomes established — after it, subsequent turns mostly execute or rationalize the doomed direction. Note also the immediately preceding healthy turn as a sanity check.
4. Decide the root cause per the taxonomy below, based on the mechanism of that earliest decisive turn, not on later symptoms.

Guard against one timing mistake above all: do not drift late. The decisive turn is where the wrong direction is SET — the wrong framing, wrong exclusion, premature narrowing, or unverified acceptance first takes hold — not the later turn where the agent states the commitment explicitly or emits the answer. That the agent could in principle still have recovered after a turn does not make that turn non-decisive. Do not default to the final answer turn when the commitment happened earlier.

## Required Output

Return exactly one JSON object. Do not use Markdown, code fences, or any text outside the JSON object.
The first non-whitespace character of your response must be `{` and the last non-whitespace character must be `}`.

JSON schema:

```json
{
  "root_cause_primary": "Search Coverage Gap | Unverified Source Reliance | Candidate Mismanagement | Constraint Neglect | Entity–Relation Misbinding | Unsupported Answer",
  "critical_step": {
    "message_index": 0
  },
  "rationale": {
    "failure_rationale": "Explain from the trajectory why this critical step is the earliest decisive error and how it makes the predicted answer unreliable."
  },
  "audit_notes": {
    "timeline_scan": [
      {"message_index": 0, "verdict": "healthy | suspicious | decisive", "note": "…"}
    ],
    "last_healthy_message_index": 0,
    "alternative_roots": ["second-choice label", "third-choice label"],
    "evidence_message_indices": [0],
    "confidence": 0.0
  }
}
```

`audit_notes` is your structured working record for the adjudicator; keep `timeline_scan` to the 4-8 most informative turns.

## Root Cause Taxonomy

`root_cause_primary` is a single-label primary cause. Choose exactly one label from the fixed taxonomy. Select the most important, earliest, and most explanatory failure mode, not a list of every problem.

1. `Search Coverage Gap`
   The model did not search for or inspect a key evidence path needed to solve the problem. Typical cases include overly narrow search direction, failure to explore obvious candidates, failure to inspect a key data source, or omission of a key evidence table or official source.

2. `Unverified Source Reliance`
   The model found an apparently relevant source but did not verify source reliability or original content. Typical cases include trusting snippets, titles, secondary sources, low-quality pages, inaccessible pages, or accepting a claim without opening a primary source.

3. `Candidate Mismanagement`
   The model failed to maintain, compare, or update candidate answers correctly. Typical cases include prematurely locking onto a partial-match candidate, discarding still-viable candidates, failing to compare multiple candidates systematically, or failing to update the answer after stronger evidence appears.

4. `Constraint Neglect`
   The model accepted a candidate that does not satisfy hard query constraints. Typical cases include ignoring contradictions, relaxing explicit conditions, rationalizing mismatches, or treating a mismatch as acceptable.

5. `Entity–Relation Misbinding`
   The model found relevant facts but bound them to the wrong entity, relation, role, time, work, person, organization, or answer slot. The core problem is that the fact was attached to the wrong target, not that no fact was found.

6. `Unsupported Answer`
   The final answer lacks support from the trajectory evidence, and no more specific search coverage, source verification, candidate management, constraint neglect, or entity-relation binding failure explains it. Typical cases include guessing, memory-based completion, or extracting an answer without an evidence chain.


## Critical Step Rule

`critical_step.message_index` should mark the earliest decisive error step: the point where the failure path becomes established and later errors are mostly continuations of that decision.

Requirements:

- `message_index` must be an existing 0-based `message_index` in the trajectory.
- `message_index` must point to an assistant message, not a tool output, user message, or system message.
- If the decisive evidence appears in a tool output, choose the assistant message that issued the tool call or the later assistant message that accepted, ignored, or misused that tool evidence.
- Never output the `message_index` of a `role="tool"` message.
- `message_index` should be the earliest key error, not the final answer by default and not a later repetition of the same mistake.
- If an assistant message makes the harmful candidate commitment, wrong exclusion, bad inference, constraint relaxation, or unverified source acceptance explicit, prefer that message.


## Rationale

`rationale.failure_rationale` explains why the selected critical step is the earliest decisive error and how it makes the `predicted_answer` wrong or unreliable.

Requirements:

- Use only the query, predicted answer, and trajectory evidence.
- You may explain which query constraints the predicted answer fails to satisfy.
- You may explain what evidence path was omitted, what candidate was wrongly accepted or discarded, or what verification was not completed.
- Do not write "the correct answer is X", "the gold answer is X", or "it should be X instead of Y".
- Do not use the hidden answer as the basis for the rationale.

Now audit the JSON input supplied by the user and return only the JSON object matching the schema.

\end{searchauditorprompt}

\subsection{Stage 2: Evidence-Grounded Adjudication}

The adjudicator receives compact replica reports, their agreement summary, the assistant outline, evidence windows, and, in the default configuration, the full trajectory.

\begin{searchauditorprompt}[label={prompt:searchauditor-adjudicator}]{Prompt 4. Evidence-Grounded Adjudicator}
You are the chief adjudicator of a SearchAuditor audit panel.

Several independent auditor replicas have each diagnosed the same failed search-agent trajectory: why it produced an unreliable or wrong `predicted_answer`, which root cause applies, and which assistant step was the earliest decisive error. Their reports disagree in part. Your job is to re-decide each field of the final diagnosis, using the replica reports as arguments and the trajectory evidence as the ground you verify against.

You will see, in the user JSON:

- `query`, `predicted_answer`
- `assistant_outline`: deterministic index of the assistant decision turns
- `replica_reports`: each replica's root cause, critical step, rationale, and structured audit notes
- `agreement_summary`: deterministic vote counts and step clustering (evidence, not a decision)
- `evidence_windows`: verbatim trajectory segments around every proposed critical step and around the final answer
- `full_trajectory` (when present): the complete trajectory

You must not rely on or infer the hidden gold answer. You will not receive the gold answer. Do not treat any replica as privileged.

## Adjudication Rules

1. **Majority vote is evidence only. A minority replica can be correct.** Never pick a label or step merely because two replicas said so; verify against the evidence windows and trajectory.
2. Re-decide **field by field**: first `root_cause_primary`, then `critical_step.message_index`. The two must be consistent — the chosen step must be where the chosen failure mechanism is actually committed.
3. For every proposed critical step, read its evidence window before judging. If all proposals are wrong and the true decisive step is elsewhere, you may override with a different assistant `message_index`, but only after locating it in the trajectory.
4. Record provenance: for each decided field, note which replica you sided with, or `adjudicator_override`.


## Critical-Step Calibration

`critical_step.message_index` marks the EARLIEST decisive error step — the point where the failure path becomes established and later errors are mostly continuations of that decision.

- Earliest establishment, not latest confirmation. If a wrong framing, wrong exclusion, premature narrowing, skipped verification, or constraint relaxation is introduced at step k and the rest of the trajectory follows from it, the decisive step is k — NOT the later step where the agent makes the commitment explicit, repeats it, or finalizes the answer.
- That recovery remained theoretically possible after step k does NOT make step k non-decisive. Almost every failed trajectory could in principle have been rescued later; the decisive step is where the wrong path was set, not the last chance to fix it.
- It is not the final answer by default: point at the answer-emission turn only when the failure is introduced there (e.g., an unsupported assertion that contradicts or ignores the agent's own verified findings).
- If the decisive information appeared in a tool output, choose the assistant message that issued the doomed call or the earliest assistant message that misused, ignored, or accepted that tool evidence — never a `role="tool"` message index.
- When replica proposals disagree, verify each proposal's evidence window; among proposals the evidence supports, prefer the EARLIEST. Never move later than an evidence-supported proposal merely because a later step states the commitment more explicitly. Choose a step outside all replica proposals only if you can cite trajectory evidence that a different assistant message established the failure path.

## Root Cause Taxonomy (fixed labels)

1. `Search Coverage Gap` — The model did not search for or inspect a key evidence path needed to solve the problem. Typical cases include overly narrow search direction, failure to explore obvious candidates, failure to inspect a key data source, or omission of a key evidence table or official source.
2. `Unverified Source Reliance` — The model found an apparently relevant source but did not verify source reliability or original content. Typical cases include trusting snippets, titles, secondary sources, low-quality pages, inaccessible pages, or accepting a claim without opening a primary source.
3. `Candidate Mismanagement` — The model failed to maintain, compare, or update candidate answers correctly. Typical cases include prematurely locking onto a partial-match candidate, discarding still-viable candidates, failing to compare multiple candidates systematically, or failing to update the answer after stronger evidence appears.
4. `Constraint Neglect` — The model accepted a candidate that does not satisfy hard query constraints. Typical cases include ignoring contradictions, relaxing explicit conditions, rationalizing mismatches, or treating a mismatch as acceptable.
5. `Entity–Relation Misbinding` — The model found relevant facts but bound them to the wrong entity, relation, role, time, work, person, organization, or answer slot. The core problem is that the fact was attached to the wrong target, not that no fact was found.
6. `Unsupported Answer` — The final answer lacks support from the trajectory evidence, and no more specific search coverage, source verification, candidate management, constraint neglect, or entity-relation binding failure explains it. Typical cases include guessing, memory-based completion, or extracting an answer without an evidence chain.

## Required Output

Return exactly one JSON object. Do not use Markdown, code fences, or any text outside the JSON object.
The first non-whitespace character of your response must be `{` and the last non-whitespace character must be `}`.

JSON schema:

```json
{
  "root_cause_primary": "one of the six labels",
  "critical_step": {
    "message_index": 0
  },
  "selected_from": {
    "root_cause": "replica id or adjudicator_override",
    "critical_step": "replica id or adjudicator_override"
  },
  "rationale": {
    "failure_rationale": "Final, self-contained explanation: why this assistant step is the earliest decisive error, citing concrete message indices and what they show, and how the failure propagates to the predicted answer."
  },
  "adjudication_note": "1-3 sentences: how the replica disagreement was resolved and which evidence settled it."
}
```

Rules for the final `failure_rationale`:

- Use only the query, predicted answer, and trajectory evidence; cite message indices you actually verified.
- Do not write "the correct answer is X" or otherwise assert the hidden answer.
- It must stand alone: a reader who has not seen the replica reports must understand the diagnosis.

Now adjudicate the JSON input supplied by the user and return only the JSON object matching the schema.

\end{searchauditorprompt}

\subsection{Stage 3: Diagnosis-Conditioned Repair Synthesis}

The final synthesizer receives the fixed diagnosis and bounded evidence windows around the adjudicated critical step and final answer, but not the full trajectory.

\begin{searchauditorprompt}[label={prompt:searchauditor-repair}]{Prompt 5. Diagnosis-Conditioned Repair Synthesizer}
You are the repair writer of a SearchAuditor audit panel.

The panel has already diagnosed a failed search-agent trajectory: the root cause and the earliest decisive assistant step are fixed and given to you, together with verbatim trajectory evidence around that step and around the final answer. Your only job is to write the final `repair_directive`: the concrete process-level instruction that, applied at the decisive step, would put this specific investigation back on a reliable path.

You will see, in the user JSON: `query`, `predicted_answer`, `final_diagnosis` (root cause, critical step index, failure rationale), and `evidence_windows`.

You must not rely on or infer the hidden gold answer. You will not receive the gold answer.

Write the repair as **2 to 4 atomic directives joined by semicolons**, in one string. Each directive must:

1. Name its concrete target in THIS case — the exact search terms or query reformulation to run, the exact source/site/table/page type to open, the exact constraint or fact to re-verify (quote the query's own wording), the exact candidate operation (reject candidate X for failing constraint Y; re-open dropped candidate line Z; enumerate all candidates satisfying condition W and compare them on the full constraint set), or the exact quantity to recompute and from what.
2. Be executable: an operator rerunning the search from the decisive step could follow it without further interpretation.
3. Stay process-level: never state, name, or numerically give the hidden target answer. For count/ranking/date/numeric questions, require recomputation or re-extraction from an authoritative source instead of giving the value. Naming a WRONG candidate that must be rejected (e.g., the predicted answer) is allowed and encouraged; naming or pointing to the presumed correct final answer is forbidden.
4. Match the diagnosed root cause: repair the mechanism (coverage, source verification, candidate management, constraint checking, entity binding, or evidence grounding), not just the symptom.

Recommended directive shapes (adapt the concrete items from this case):

- "Reject <predicted candidate> because <specific constraint from the query> is violated/unverified per <message evidence>; rebuild the candidate set by searching <exact decisive clue terms combined>, keeping every candidate that matches <clue A> and <clue B>."
- "Return to the search at message <k>: instead of <the framing the agent locked into>, run queries combining <exact term 1>, <exact term 2>, <exact term 3>, and open the <official/primary source type> that lists <the needed table/fact>."
- "Before committing, build a constraint table over <list the query's hard constraints verbatim>, verify each against opened primary sources rather than snippets, and commit only if every hard constraint has supporting evidence."
- "Recompute <the requested count/rank/metric> directly from <the authoritative source/table to open>, aligning <the definition the query uses>, instead of relying on <the snippet/secondary claim the agent used>."

Silent self-check before answering (do not output it): for each directive verify (a) it names at least one concrete case item (term, source, constraint, candidate, metric); (b) it is executable at or after the decisive step; (c) it reveals no hidden answer value; (d) together the directives address the diagnosed root cause. Rewrite any directive that fails.

## Required Output

Return exactly one JSON object. Do not use Markdown, code fences, or any text outside the JSON object.
The first non-whitespace character of your response must be `{` and the last non-whitespace character must be `}`.

JSON schema:

```json
{
  "repair_directive": "directive 1; directive 2; directive 3"
}
```

Now write the repair for the JSON input supplied by the user and return only the JSON object matching the schema.

\end{searchauditorprompt}


\end{document}